%% file: main_cameraready.tex
\documentclass{article}

\usepackage[dblblindworkshop,nonatbib,final]{neurips_2026}
\usepackage[numbers,sort&compress]{natbib}

\usepackage[utf8]{inputenc}
\usepackage[T1]{fontenc}

\usepackage{amsmath}
\usepackage{amssymb}
\usepackage{graphicx}
\usepackage{booktabs}
\usepackage{multirow}
\usepackage{threeparttable}
\usepackage{longtable}
\usepackage[table]{xcolor}   %
\usepackage{float}
\usepackage{algorithm}
\usepackage[noend]{algpseudocode}
\usepackage{subcaption}
\usepackage{microtype}
\usepackage{xspace}

\usepackage{hyperref}
\usepackage{cleveref}   %

\hypersetup{
  pdftitle={Class-wise LoRA Anchoring for Continual Post-Training of Open-Vocabulary Detectors},
  pdfauthor={Deokyun Kim, Giyoung Lee, Yookyung Kim, Jihun Cha},
  pdfsubject={NeurIPS 2026 Workshop on Continual Learning in the Era of Foundation Models and Embodied Agents},
  pdfkeywords={continual learning, open-vocabulary object detection, parameter-efficient fine-tuning, LoRA, catastrophic forgetting},
}

\usepackage{accvabbrv}

\graphicspath{{figures/}}

\usepackage[font=footnotesize,labelfont=bf]{caption}
\usepackage{enumitem}   %

\newcommand{\rev}[1]{#1}

\newcommand{\hi}[1]{\cellcolor{yellow!40}#1}
\newcommand{\hipos}[1]{\cellcolor{cyan!30}#1}
\newcommand{\tup}[1]{#1\,\textcolor{green!55!black}{\ensuremath{\blacktriangle}}}
\newcommand{\tdn}[1]{#1\,\textcolor{red!75!black}{\ensuremath{\blacktriangledown}}}

\workshoptitle{Continual Learning in the Era of Foundation Models and Embodied Agents}

\newcommand{\codeavailability}{Source code is available at
  \url{https://github.com/DeokyunKim/Class-wise_LoRA_Anchoring_Incremental_Learning_VLOD}.}

\newcommand{\codeanswer}{\answerYes}
\newcommand{\codejustification}{Our code, with training and evaluation scripts, is
  released at \url{https://github.com/DeokyunKim/Class-wise_LoRA_Anchoring_Incremental_Learning_VLOD} and linked from the abstract.}

\title{Class-wise LoRA Anchoring for Continual Post-Training of Open-Vocabulary Detectors}

\author{%
  Deokyun Kim \quad
  Giyoung Lee \quad
  Yookyung Kim \quad
  Jihun Cha \\[2pt]
  Autonomous UAV Research Section \\
  Electronics and Telecommunications Research Institute, Republic of Korea \\
  \texttt{\{deokyunkim, giyoung, yk.kim, jihun\}@etri.re.kr} \\
}

\begin{document}

\maketitle

\input{sections/abstract.tex}

\input{sections/01_introduction.tex}
\input{sections/02_related_work.tex}
\input{sections/03_method.tex}

\input{sections/04_experiments.tex}
\input{sections/05_conclusion.tex}

\label{endofmaintext}

\begin{ack}
This work was supported by Institute of Information \& communications
Technology Planning \& Evaluation (IITP) grant funded by the Korea government
(MSIT) (No.\ RS-2022-II220021, Development of Core Technologies for Autonomous
Searching Drones).
\end{ack}

\bibliographystyle{plainnat}
\bibliography{main}

\clearpage
\appendix
\phantomsection
\section*{Appendix}
\input{sections/06_appendix.tex}

\clearpage
\input{checklist.tex}

\end{document}

%% file: sections/abstract.tex
\begin{abstract}
  Continual post-training of a vision-language foundation model on a task stream erodes the zero-shot ability acquired in pretraining, a phenomenon known as catastrophic forgetting. We study this in open-vocabulary object detection, whose detectors start from strong zero-shot ability. Parameter-efficient methods mitigate this by training lightweight modules such as LoRA over a frozen backbone, and recent systems organize them into a reusable library of class-specific modules. How the modules in such a library should be optimized as the task sequence progresses is largely unexamined. We propose Class-wise LoRA Anchoring (CLA), a library-aware L2 anchor that keeps each class module close to its task-start state, anchoring a reused module to its consolidated prior and a new module to its low-correlated random initialization. On full-shot ODinW-13, CLA raises post-adaptation zero-shot COCO above the state-of-the-art class-wise library, DitHub, and above the unadapted detector itself, so adaptation stops trading generality for task accuracy, which stays comparable. The gain concentrates on categories never seen during adaptation and comes with a markedly less correlated class library, which makes training-free class unlearning more effective. CLA is most effective in class-incremental regimes that add new categories, and reveals that a retention-adaptation trade-off remains in domain-incremental settings. \codeavailability
\end{abstract}

%% file: sections/01_introduction.tex
\section{Introduction}
\label{sec:intro}
Open-vocabulary object detection, represented by GLIP~\cite{li2022glip} and Grounding DINO~\cite{liu2024grounding}, localizes objects specified by natural-language queries beyond a predefined category set. Large-scale vision-language pretraining endows them with strong zero-shot detection capability. In practice, downstream tasks frequently involve rare categories, specialized imagery, or annotation distributions underrepresented during pretraining. Sequentially adapting such a detector to these tasks can bias the resulting model toward the adaptation stream and erode this capability, a phenomenon known as catastrophic forgetting. Parameter-efficient fine-tuning mitigates catastrophic forgetting by training lightweight modules, such as low-rank adapters~\cite{hu2022lora}, while keeping the pretrained backbone weights frozen. Freezing the backbone alone, however, does not preserve zero-shot detection capability. To keep performing the downstream tasks, the adapters stay active in the inference path for every query; and because they are fitted to those tasks, they shift the feature distribution of the backbone toward the downstream domain even for open-vocabulary inputs the model was never adapted on. The zero-shot detection capability therefore degrades despite the frozen backbone, and effective adaptation requires the adapters themselves to retain this capability, not merely to fit the downstream tasks.

\rev{Recent work, ZiRa~\cite{deng2024zira},} formalizes Incremental Vision-Language Object Detection (IVLOD), adapting a pretrained detector to specialized downstream domains while preserving its zero-shot generalization, and balances stability and plasticity through a reparameterizable dual-branch adapter and a zero-interference loss. DitHub~\cite{cappellino2025dithub}, in contrast, organizes class-specific LoRA modules into a reusable library, assigning a per-class factor $A$ to each object class while sharing $B$ across classes, and attains state-of-the-art performance on ODinW benchmarks. Such libraries let class-level knowledge be fetched, updated, and merged across tasks, but their optimization remains underexplored: revisited modules are still driven only by the detection loss of the current task, without explicit constraints on their prior library state or on their relation to the other class-specific $A$ modules in the same layer.

We propose Class-wise LoRA Anchoring (CLA), a library-aware objective for class-wise LoRA libraries, instantiated on DitHub. At each task, CLA snapshots the current state of the module library and anchors each trainable module to that snapshot with an L2 distance penalty. For a recurring class, the reference is its consolidated prior-task module; for a new class, it is the module's low-correlated random initialization. The anchor therefore keeps reused modules close to their consolidated state and weakly stabilizes newly initialized modules, and its strength $\gamma_{\mathrm{a}}$ trades task adaptation against zero-shot retention.

Experiments on the ODinW benchmarks~\cite{li2022elevater} show that CLA improves post-adaptation zero-shot COCO AP (ZCOCO) under both protocols, while the trade-off against the average detection AP over the downstream tasks (Avg) depends on how the tasks share classes. On the mostly class-disjoint ODinW-13~\cite{cappellino2025dithub}, CLA raises ZCOCO while keeping Avg comparable to DitHub, and a class-wise COCO decomposition shows that CLA improves the categories seen during adaptation and those never seen together, whereas DitHub improves the seen ones and forgets the rest. On class-overlapped ODinW-O, a domain-incremental regime outside our class-incremental focus, CLA again prioritizes zero-shot retention, at a modest Avg cost.

Our contributions are summarized as follows:
\begin{itemize}[nosep,leftmargin=1.3em]
    \item We propose \textbf{Class-wise LoRA Anchoring (CLA)}, an L2 anchor that ties each class-specific LoRA module in the DitHub library to its task-start state: a reused module to the consolidated prior, and a new module to the low-correlated random initialization.

    \item We demonstrate that CLA improves the retention-adaptation balance: on ODinW-13 it raises zero-shot COCO above the unadapted Grounding DINO while keeping average task AP comparable to DitHub, and on ODinW-O it attains the strongest zero-shot retention among adapted methods.

    \item \rev{Our analyses locate the retention gain of CLA: it concentrates on COCO categories never seen during adaptation and on a low-correlated class-specific $A$ library, which also makes training-free class unlearning more effective; a selective ablation shows the gain is largely lost without the new-class anchor, regressing the library toward DitHub despite a smaller perturbation.}
\end{itemize}

%% file: sections/02_related_work.tex
\section{Related Work}

\paragraph{Open-Vocabulary Object Detection.}
Open-vocabulary object detection localizes objects specified by natural-language queries rather than a fixed category set. Vision-language detectors~\cite{li2022glip, liu2024grounding, cheng2024yoloworld, minderer2023owlv2, yao2022detclip} learn region-text alignment from large-scale supervision, enabling zero-shot localization beyond predefined categories. Such detectors are routinely adapted to specialized settings~\cite{pan2025locate, zhang2024opensight, wasim2024videogrounding}, but adaptation is rarely one-time: \rev{new categories and conditions keep arriving, so it becomes incremental and must preserve previously acquired capability, often without earlier-task data for replay.}

\paragraph{Parameter-Efficient Fine-Tuning and Forgetting Mitigation.}
\rev{Parameter-efficient Fine-Tuning (PEFT)} specializes large-scale vision-language models by updating lightweight modules while keeping most pretrained weights fixed, through low-rank updates~\cite{hu2022lora}, adapters~\cite{houlsby2019adapter, gao2024clipadapter, sung2022vladapter}, or prompt, selective, and sparse-expert tuning~\cite{khattak2023maple, li2024clipfit, lu2025dynamicdino}. PEFT, however, does not by itself preserve zero-shot detection capability, as learned adapter updates remain active during inference and can bias the detector toward \rev{the downstream tasks it is adapted on}.

Meanwhile, continual learning adapts a model to a sequence of tasks while limiting interference with previously acquired knowledge. Regularization-based methods such as EWC~\cite{kirkpatrick2017ewc}, SI~\cite{zenke2017si}, and MAS~\cite{aljundi2018mas} constrain changes to parameters important for earlier tasks; distillation-based methods such as LwF~\cite{li2018lwf} and iDETR~\cite{dong2023idetr} transfer knowledge from a previous model to the current one; and exemplar-replay-based methods such as OW-DETR~\cite{gupta2022owdetr} and CL-DETR~\cite{liu2023cldetr} rehearse stored samples of earlier tasks, all mitigating forgetting when learning new classes. \rev{These were largely developed for closed-set classification or standard detection; for vision-language detectors the central additional difficulty is retaining the zero-shot detection capability inherited from pretraining.}

\paragraph{Incremental Open-Vocabulary Object Detection.}

\rev{For open-vocabulary detection, ZiRa~\cite{deng2024zira} formalizes Incremental Vision-Language Object Detection (IVLOD) and adapts Grounding DINO with a reparameterizable dual-branch adapter and a zero-interference loss penalizing the adapter output norm, so that a smaller perturbation of the pretrained detector preserves its zero-shot ability. DitHub~\cite{cappellino2025dithub} instead casts adaptation as version control over a class-wise LoRA library: it assigns a per-class low-rank factor $A$ to each object class, shares one $B$ across classes, and fetches, updates, and merges the relevant class modules as tasks arrive, reaching state-of-the-art results on ODinW. MR-GDINO~\cite{dong2024mrgdino} instead retains zero-shot ability through concept and interaction memory.}

The remaining gap is how the reusable class-wise modules of such libraries should be optimized over a task sequence, as each fetched module is still trained mainly on the incoming downstream task data under the detection loss. CLA addresses this by anchoring each module to its task-start state: a reused module to its consolidated prior, and a new module to its low-correlated random initialization. \rev{L2-SP~\cite{li2018l2sp} anchors all adapted parameters to a fixed pretrained reference, whereas CLA anchors only the active library components and refreshes the reference at every task boundary, making it a library-aware constraint rather than a global weight regularizer.} Orthogonal-LoRA ideas appear in language-model continual learning~\cite{wang2023olora} and domain generalization~\cite{hu2024pego}, but add an explicit orthogonality term; CLA instead anchors each new module to its low-correlated initialization, which we find empirically keeps the class-specific modules less correlated.

%% file: sections/03_method.tex
\section{Methodology}

\subsection{Overview}

\begin{figure}[!ht]
  \centering
  \includegraphics[width=1.0\linewidth]{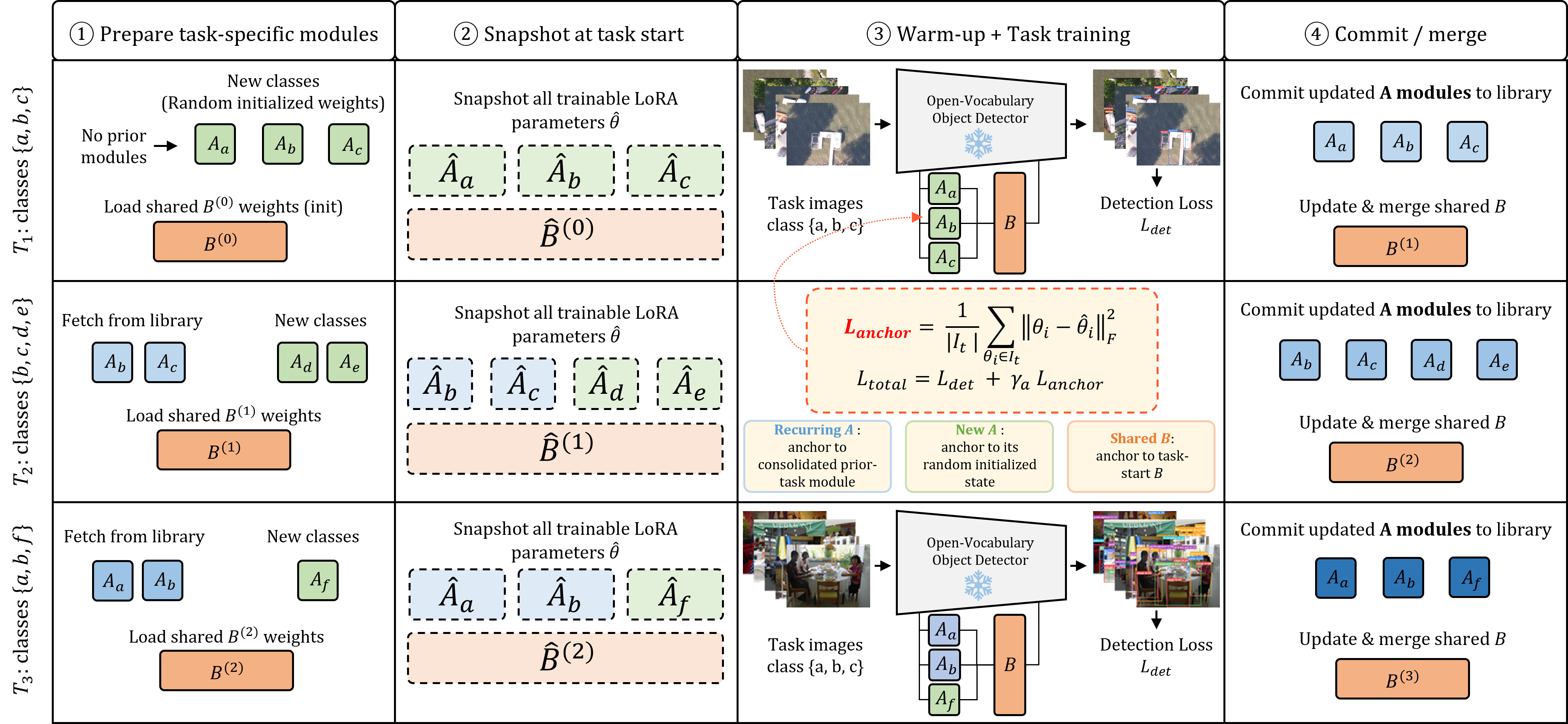}
  \caption{Overview of Class-wise LoRA Anchoring (CLA). CLA snapshots the task-start LoRA state and anchors each trainable module to that snapshot: a recurring class-specific $A$ to the consolidated prior-task module, a newly introduced class-specific $A$ to its low-correlated random initialization, and the shared $B$ to the task-start state.}
  \label{fig:cla_overview}
\end{figure}

We build upon DitHub~\cite{cappellino2025dithub}, a parameter-efficient fine-tuning method for incremental open-vocabulary detection that stores LoRA-based adaptation~\cite{hu2022lora} as a class-wise library of $A$ matrices and shared $B$ matrices. DitHub accumulates class modules efficiently, but each module is optimized by the general object detection loss alone, without accounting for the knowledge consolidated in earlier tasks. The active adapter can then weaken the zero-shot detection capability inherited from vision-language pretraining.

We propose \textbf{Class-wise LoRA Anchoring (CLA)}, shown in \Cref{fig:cla_overview}: a library-aware training term, an \emph{L2 anchor} that, at each task, keeps every trainable LoRA module close to its task-start state. For a recurring class, this reference is its consolidated prior-task module; for a class appearing for the first time, it is the low-correlated random initialization of the module. CLA is thus a task-start LoRA regularizer: it anchors reused class-specific modules to their consolidated task-start states and regularizes the shared LoRA parameters to reduce task-induced drift, while for a newly initialized module the anchor acts as a weak stabilizer rather than a memory-preservation term. A strength $\gamma_{\mathrm{a}}$ trades task adaptation against zero-shot retention.

\subsection{Preliminaries: Class-specific LoRA Modules in DitHub}
\label{sec:prelim}

DitHub~\cite{cappellino2025dithub} applies LoRA-based adaptation~\cite{hu2022lora} to a pretrained open-vocabulary detector such as Grounding DINO~\cite{liu2024grounding}. For each adaptable layer, LoRA adds a low-rank update to the frozen weight $W$, giving the adapted weight $W' = W + BA$ with trainable matrices $A$ and $B$. Updating only these matrices adapts the detector efficiently while its pretrained weights stay fixed.

DitHub assigns a class-specific $A$ matrix $A_c$ to each object class and stores them in a Modules Library, while sharing $B$ across classes, so that the effective low-rank update for class $c$ is $B A_c$. At task $t$, it fetches existing class-specific $A$ modules for \rev{recurring object classes in the current task}, gives new-class modules a low-correlated random initialization, trains the fetched $A$ modules and shared $B$ matrices on the current \rev{task}, and merges the updated class modules back into the library. This design learns the object categories appearing across the downstream tasks, but the modules are not explicitly tied to the LoRA state consolidated before the task. CLA instead ties each module to the LoRA state it had at the start of the task.

\subsection{Class-wise LoRA Anchoring}
\label{sec:cla}

CLA makes the DitHub objective library-aware: each trainable LoRA module is optimized relative to the state the library already holds for it at the start of the task, rather than by the detection loss alone. An L2 anchor enforces this reference, so the reused class-specific and shared modules adapt to the downstream detection task while their task-induced drift stays bounded.

\paragraph{Class-wise anchoring.}
Let $\Theta_t$ denote the trainable LoRA parameters during task $t$, including fetched and newly initialized class-specific $A$ matrices and shared $B$ matrices. At the start of every task, after the modules are fetched and initialized and before any gradient update, CLA stores a detached snapshot $\hat{\theta}_i$ of each trainable module $\theta_i$ as its anchor target. Let $\mathcal{R}_t$ be the reused class modules and the shared $B$, and $\mathcal{N}_t$ the modules appearing for the first time, with $\mathcal{I}_t=\mathcal{R}_t\cup\mathcal{N}_t$; the anchor target is
\begin{equation}
  \hat{\theta}_i =
  \begin{cases}
    \bar{\theta}_i, & \theta_i \in \mathcal{R}_t,\\[2pt]
    \theta_i^{\mathrm{init}}, & \theta_i \in \mathcal{N}_t,
  \end{cases}
\end{equation}
with $\bar{\theta}_i$ the consolidated prior state from an earlier task and $\theta_i^{\mathrm{init}}$ the random initialization. Applied at every task, including the first, the anchor penalizes each module's L2 deviation from its snapshot,
\begin{equation}
    \mathcal{L}_{\mathrm{anchor}}
    =
    \frac{1}{|\mathcal{I}_t|}
    \sum_{\theta_i \in \mathcal{I}_t}
    \left\| \theta_i - \hat{\theta}_i \right\|_F^2 .
\end{equation}
In our main configuration, $\mathcal{I}_t$ includes all class-specific $A$ matrices and shared $B$ matrices. \rev{Because class-specific modules are stored in and fetched from the library, absent class modules are not updated during unrelated tasks; CLA therefore mainly reduces the drift of the modules actively reused or shared in the current task.}

For a recurring class the target is a consolidated prior state, and the penalty preserves the knowledge acquired in earlier tasks. For a newly introduced class the target carries no task knowledge, and the penalty instead bounds the displacement of the module from its initialization. Because independent random initializations are near-orthogonal in the high-dimensional LoRA space, bounding this displacement keeps the pairwise correlation between class modules low and limits the perturbation $\|B A_c\|$ that the library applies to the frozen detector. CLA adds the anchor to the task loss:
\begin{equation}
    \mathcal{L}_{\mathrm{total}}
    =
    \mathcal{L}_{\mathrm{det}}
    +
    \gamma_{\mathrm{a}} \mathcal{L}_{\mathrm{anchor}},
\end{equation}
where \rev{$\mathcal{L}_{\mathrm{det}}$ is the original DitHub object detection loss} and $\gamma_{\mathrm{a}}$ controls the anchor strength. Because CLA only adds a training-time loss, the inference path is identical to that of DitHub, with no additional overhead.

\paragraph{Training algorithm.}
Algorithm~\ref{alg:cla} summarizes one task of CLA training. CLA snapshots the LoRA state once before task training, refreshes the snapshot at every task, and adds $\mathcal{L}_{\mathrm{anchor}}$ through the same backward pass as the detection loss.

\algrenewcommand\algorithmicrequire{\textbf{Input:}}
\algrenewcommand\algorithmicensure{\textbf{Output:}}
\newlength{\algcommentwidth}
\begin{algorithm}[!ht]
\caption{CLA training for one incremental task.}
\label{alg:cla}
\footnotesize
\settowidth{\algcommentwidth}{$\triangleright$~fetch reused $A$; random-init new $A$; load $B$}
\algrenewcommand\algorithmiccomment[1]{\hfill\makebox[\algcommentwidth][l]{$\triangleright$~#1}}
\begin{algorithmic}[1]
\Require Model $f$; task $t$; data $\mathcal{D}_t$; library $\mathcal{A}$; anchor target $T$; weight $\gamma_{\mathrm{a}}$; warm-up iter $t_{\mathrm{w}}$; coefficients $\lambda_a,\lambda_b$.
\Ensure Updated library $\mathcal{A}$.
\State Prepare LoRA modules. \Comment{fetch reused $A$; random-init new $A$; load $B$}
\State $\hat{\Theta}\gets \mathrm{Snapshot}(f,T)$ if $\gamma_{\mathrm{a}}{>}0$, else $\emptyset$. \Comment{reused $A\!\to$ prior, new $A\!\to$ random init}
\For{iteration $i$, mini-batch $x \sim \mathcal{D}_t$}
    \If{$i = t_{\mathrm{w}}$} \Comment{class-agnostic warm-up~\cite{cappellino2025dithub}}
        \State $A_c \gets \lambda_a A_c^{\mathrm{warm}} + (1{-}\lambda_a) A_c$
        \State $B \gets \lambda_b B^{\mathrm{warm}} + (1{-}\lambda_b) B$
    \EndIf
    \State $\mathcal{L}_{\mathrm{det}} \gets \mathrm{DetectionLoss}(f,x)$
    \State $\mathcal{L}_{\mathrm{anchor}} \gets \mathrm{L2Anchor}(f,\hat{\Theta},T)$ \Comment{anchor $A$, $B$ to $\hat{\Theta}$}
    \State Update by minimizing $\mathcal{L}_{\mathrm{total}}=\mathcal{L}_{\mathrm{det}}+\gamma_{\mathrm{a}}\mathcal{L}_{\mathrm{anchor}}$.
\EndFor
\State Commit updated $A$ to $\mathcal{A}$; merge shared $B$ across tasks.
\end{algorithmic}
\end{algorithm}

%% file: sections/04_experiments.tex
\section{Experiments}

\subsection{Setup}
\label{sec:evaluation}

\paragraph{Setup and benchmarks.}
Following DitHub~\cite{cappellino2025dithub}, we use Grounding DINO~\cite{liu2024grounding} as the base open-vocabulary detector and adopt the same class-wise LoRA Modules Library, to which \rev{CLA adds a class-wise L2 anchor}. We evaluate on the two incremental open-vocabulary detection protocols DitHub introduces, both derived from the ODinW~\cite{li2022elevater} suite. \textit{ODinW-13}~\cite{li2022glip} consists of 13 datasets: Aerial Maritime Drone (Ae), Aquarium (Aq), Cottontail Rabbits (Co), EgoHands (Eg), North America Mushrooms (Mu), Packages (Pa), Pascal VOC (Pv), Pistols (Pi), Pothole (Po), Raccoon (Ra), Shellfish (Sh), Thermal Dogs and People (Th), and Vehicles (Ve); it covers heterogeneous domains with mostly disjoint classes. \textit{ODinW-O} is a 6-task subset for class-overlap forgetting: Aerial Maritime Drone (Ae), Hard Hat Workers (Hw), Pascal VOC (Pv), Self-driving Car (Sd), Thermal Dogs and People (Th), and Vehicles (Ve).

\paragraph{Metrics.}
We report \textbf{ZCOCO}, zero-shot AP on COCO~\cite{lin2014coco} val2017 after the full incremental training, used as a proxy for retained open-vocabulary detection capability; and \textbf{Avg}, mean per-task AP at the end of the incremental sequence.
All AP values use IoU~$\in[0.50,0.95]$. Results from our runs are reported over 3 random task-order seeds, with standard deviations shown when available.

\paragraph{Baselines and implementation.}
We compare with the baselines DitHub reports: TFA~\cite{wang2020tfa}, iDETR~\cite{dong2023idetr}, Adapter Tuning~\cite{houlsby2019adapter}, OW-DETR~\cite{gupta2022owdetr}, CL-DETR~\cite{liu2023cldetr}, ZiRa~\cite{deng2024zira}, DitHub, and the zero-shot Grounding DINO reference. External baseline and DitHub rows in the main tables are quoted from DitHub; CLA rows and the diagnostic analyses use our runs, with seed-matched DitHub reproductions. Our implementation is based on DitHub, except that we use $\lambda_a{=}0.3,~\lambda_b{=}0.7$ for both benchmarks rather than the $\lambda_a{=}0.1$ used by DitHub for reappearing classes on ODinW-O. We set $\gamma_{\mathrm{a}}{=}1.0$ on ODinW-13 and $\gamma_{\mathrm{a}}{=}0.3$ on ODinW-O, and train for $46{,}700$ and $42{,}600$ iterations per seed, respectively; further implementation details are given in Appendix~\ref{sec:supp_impl} and a sweep of $\gamma_{\mathrm{a}}$ in Appendix~\ref{sec:supp_gamma}.

\subsection{Benchmark Results}
\label{sec:results}

\paragraph{ODinW-13.}

\begin{table}[t]
  \caption{Per-task and aggregate mAP on \textbf{ODinW-13} at $0/1/5/10/$Full shots. ZCOCO is the post-adaptation zero-shot COCO AP and Avg the per-task mean. DitHub and external baselines are quoted from~\cite{cappellino2025dithub}.}
  \label{tab:odinw13_main}
  \centering
  \scriptsize
  \setlength{\tabcolsep}{1.5pt}
  \renewcommand{\arraystretch}{0.9}
  \begin{tabular*}{\textwidth}{@{\extracolsep{\fill}}llcc|ccccccccccccc@{}}
    \toprule
    Shots & Method & ZCOCO $\uparrow$ & Avg $\uparrow$ & Ae & Aq & Co & Eg & Mu & Pa & Pv & Pi & Po & Ra & Sh & Th & Ve \\
    \midrule
    0 & G-DINO~\cite{liu2024grounding} & 47.41 & 46.80 & 19.11 & 20.82 & 64.75 & 59.98 & 25.34 & 56.27 & 54.80 & 65.94 & 22.13 & 62.02 & 32.85 & 70.38 & 57.07 \\
    \midrule
    \multirow{6}{*}{1} & TFA~\cite{wang2020tfa} & 18.84 & 39.50 & 18.25 & 15.81 & 63.90 & 50.79 & 28.47 & 50.37 & 29.49 & 59.16 & 21.90 & 50.67 & 19.86 & 60.85 & 43.97 \\
     & iDETR~\cite{dong2023idetr} & 44.61 & \textbf{49.82} & \textbf{22.81} & 23.24 & \textbf{69.75} & 61.43 & 31.73 & 56.27 & 55.40 & 62.44 & \textbf{28.45} & \textbf{60.33} & \textbf{43.33} & \textbf{73.64} & 58.84 \\
     & AT~\cite{houlsby2019adapter} & 44.11 & 46.23 & 21.55 & 23.62 & 66.60 & 58.96 & 27.68 & 53.97 & 54.58 & 62.47 & 26.94 & 53.17 & 20.37 & 70.31 & 60.71 \\
     & ZiRa~\cite{deng2024zira} & 46.44 & 48.56 & 20.34 & 19.64 & 69.47 & 60.00 & 30.09 & 56.27 & 58.12 & \textbf{65.13} & 25.83 & 55.69 & 39.50 & 72.59 & 58.55 \\
     & DitHub~\cite{cappellino2025dithub} & 46.40 & 49.19 & 21.33 & \textbf{24.34} & 69.23 & \textbf{61.99} & \textbf{42.10} & \textbf{57.08} & \textbf{58.85} & 55.24 & 24.62 & 57.58 & 33.56 & 70.99 & 62.50 \\
     & \textbf{CLA (Ours)} & \textbf{46.74} & 48.03 & 19.84 & 22.54 & 67.56 & 61.69 & 35.10 & 55.26 & 58.45 & 55.88 & 24.55 & 58.57 & 31.90 & 70.02 & \textbf{63.08} \\
    \midrule
    \multirow{6}{*}{5} & TFA~\cite{wang2020tfa} & 26.43 & 45.76 & 21.92 & 22.30 & 67.40 & 60.72 & 30.63 & 53.56 & 46.80 & 63.60 & 26.88 & 56.26 & 28.00 & 64.28 & 52.49 \\
     & iDETR~\cite{dong2023idetr} & 43.51 & 51.65 & 25.69 & 25.53 & 70.42 & 62.98 & 49.98 & 50.54 & 54.85 & 64.80 & \textbf{33.24} & 57.64 & \textbf{42.36} & \textbf{76.51} & 56.92 \\
     & AT~\cite{houlsby2019adapter} & 43.67 & 47.16 & 14.63 & 24.97 & 66.56 & \textbf{64.19} & 38.85 & 42.03 & 55.49 & \textbf{65.20} & 27.48 & 52.68 & 32.21 & 71.23 & 57.54 \\
     & ZiRa~\cite{deng2024zira} & 45.41 & 51.77 & 24.22 & 26.40 & 71.34 & 63.96 & 49.54 & \textbf{60.51} & 60.11 & 64.39 & 32.20 & 56.31 & 40.61 & 69.16 & 54.30 \\
     & DitHub~\cite{cappellino2025dithub} & 45.86 & 52.85 & 26.84 & 28.91 & 68.46 & 60.25 & \textbf{51.93} & 55.10 & 60.43 & 59.44 & 33.18 & \textbf{68.12} & 38.05 & 75.04 & 61.34 \\
     & \textbf{CLA (Ours)} & \textbf{46.57} & \textbf{52.97} & \textbf{27.15} & \textbf{29.96} & \textbf{72.87} & 60.07 & 50.96 & 58.34 & \textbf{60.65} & 59.86 & 32.86 & 67.23 & 32.81 & 73.99 & \textbf{61.87} \\
    \midrule
    \multirow{6}{*}{10} & TFA~\cite{wang2020tfa} & 34.35 & 46.61 & 21.17 & 22.16 & 66.82 & 60.63 & 32.35 & 50.15 & 55.54 & 64.98 & 27.59 & 57.40 & 28.14 & 66.82 & 52.11 \\
     & iDETR~\cite{dong2023idetr} & 43.54 & 53.29 & 25.39 & 27.70 & 65.62 & \textbf{67.58} & 47.99 & 60.20 & 56.32 & 63.93 & 35.18 & 59.37 & \textbf{53.63} & 74.70 & 55.12 \\
     & AT~\cite{houlsby2019adapter} & 43.06 & 47.34 & 18.73 & 25.42 & 69.77 & 66.34 & 35.84 & 48.25 & 53.39 & 64.07 & 28.89 & 50.33 & 32.50 & 66.57 & 55.37 \\
     & ZiRa~\cite{deng2024zira} & 46.14 & 53.20 & 25.56 & 25.96 & 71.72 & 65.28 & 48.22 & \textbf{62.81} & 61.24 & \textbf{65.47} & 35.58 & 55.73 & 46.86 & 70.97 & 56.14 \\
     & DitHub~\cite{cappellino2025dithub} & 46.54 & \textbf{54.43} & 26.19 & \textbf{30.28} & 68.11 & 65.73 & \textbf{52.82} & 56.82 & \textbf{63.36} & 61.85 & \textbf{36.68} & 66.82 & 41.46 & \textbf{76.00} & 61.48 \\
     & \textbf{CLA (Ours)} & \textbf{46.92} & 54.41 & \textbf{26.71} & 30.21 & \textbf{72.41} & 63.18 & 52.57 & 57.63 & 61.80 & 59.51 & 36.57 & \textbf{69.07} & 39.43 & 75.92 & \textbf{62.37} \\
    \midrule
    \multirow{8}{*}{Full} & TFA~\cite{wang2020tfa} & 30.97 & 47.93 & 23.80 & 30.65 & 67.21 & 61.77 & 30.52 & 50.23 & 47.73 & 60.91 & 29.25 & 61.72 & 31.42 & 66.23 & 61.61 \\
     & iDETR~\cite{dong2023idetr} & 37.32 & 58.71 & 32.64 & 46.65 & 70.99 & 68.56 & \textbf{55.32} & 58.88 & 64.48 & 71.01 & 50.33 & 63.30 & 39.19 & 77.12 & 64.80 \\
     & AT~\cite{houlsby2019adapter} & 42.30 & 51.14 & 23.62 & 39.90 & \textbf{72.32} & 65.51 & 31.47 & 50.48 & 60.51 & 66.07 & 39.09 & 53.50 & 34.04 & 68.07 & 60.23 \\
     & OW-DETR~\cite{gupta2022owdetr} & 31.22 & 55.58 & 28.46 & 43.78 & 70.54 & 67.78 & 43.84 & 56.75 & 63.13 & 69.51 & 45.16 & 58.99 & 36.99 & 74.42 & 63.20 \\
     & CL-DETR~\cite{liu2023cldetr} & 32.15 & 57.26 & 29.35 & 45.15 & 71.94 & \textbf{69.90} & 45.21 & 58.52 & 65.10 & \textbf{71.68} & 46.58 & 60.83 & 38.14 & 76.74 & 65.18 \\
     & ZiRa~\cite{deng2024zira} & 46.26 & 57.98 & 31.76 & 47.35 & 71.77 & 64.74 & 46.53 & 62.66 & 66.39 & 71.00 & 48.48 & 63.03 & 41.44 & 76.13 & 62.44 \\
     & DitHub~\cite{cappellino2025dithub} & 47.01 & 62.19 & 34.62 & \textbf{50.65} & 70.46 & 68.56 & 49.28 & 65.57 & \textbf{69.58} & 71.10 & \textbf{56.65} & 70.88 & \textbf{52.82} & \textbf{79.30} & \textbf{68.18} \\
     & \textbf{CLA (Ours)} & \textbf{48.28} & \textbf{62.26} & \textbf{37.40} & 49.29 & 71.61 & 69.67 & 53.43 & \textbf{70.93} & 63.50 & 70.94 & 54.61 & \textbf{71.30} & 50.80 & 78.64 & 67.21 \\
    \bottomrule
  
  \end{tabular*}
\end{table}

\begin{figure}[t]
  \centering
  \includegraphics[width=0.93\linewidth]{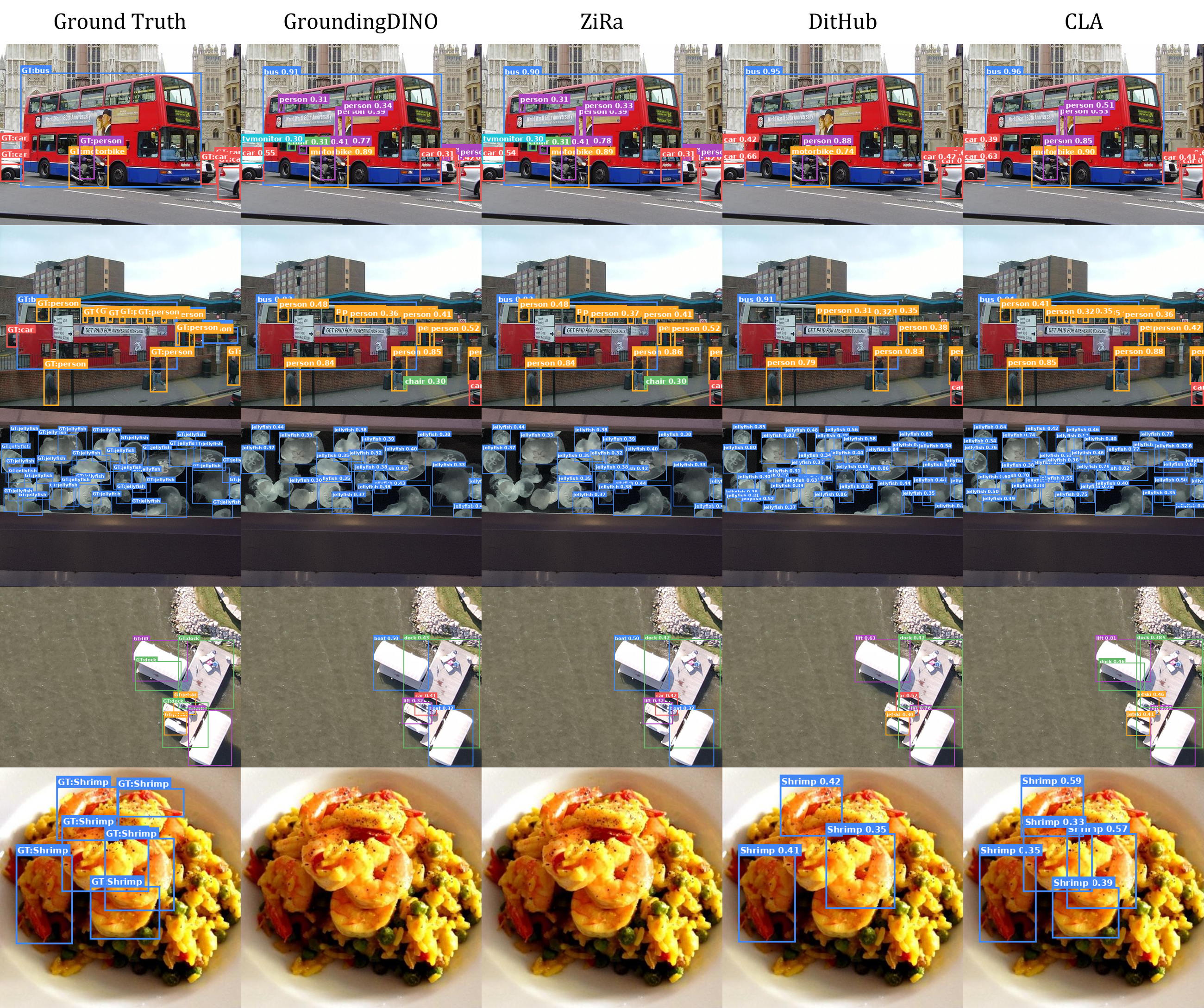}
  \caption{Qualitative results on ODinW-13. Each row is a different task; columns show, from left to right, the ground truth, Grounding DINO~\cite{liu2024grounding}, ZiRa~\cite{deng2024zira}, DitHub~\cite{cappellino2025dithub}, and CLA (Ours). Predicted boxes are shown with their confidence scores.}
  \label{fig:qualitative}
\end{figure}

Table~\ref{tab:odinw13_main} reports aggregate ZCOCO and Avg together with per-task AP across the zero-shot, few-shot, and full-shot scenarios. \rev{CLA attains the best ZCOCO in every shot regime, improving zero-shot retention over DitHub. On Avg, CLA surpasses DitHub at the $5$-shot and full-shot settings, is comparable at $10$-shot, and trails only at $1$-shot. This $1$-shot behavior is consistent with the anchoring objective: with only one annotated example per class, regularizing each module toward its prior state constrains the plasticity available in this extreme low-shot regime, yielding a small Avg reduction in exchange for stronger zero-shot retention. A particularly notable result emerges in the full-shot setting: after sequentially adapting to all $13$ tasks, the post-adaptation zero-shot COCO of CLA exceeds even the unadapted Grounding DINO by $0.87$~AP, so CLA does not merely prevent forgetting but \emph{improves} the zero-shot COCO of the base model; we further analyze this gain in Section~\ref{sec:analysis}. Figure~\ref{fig:qualitative} compares qualitative detections across several ODinW-13 tasks.}

\begin{table}[t]
  \caption{Per-task and aggregate mAP on \textbf{ODinW-O}. ZCOCO is the post-adaptation zero-shot COCO AP and Avg the per-task mean; G-DINO is the unadapted zero-shot reference.}
  \label{tab:odinwo_full}
  \centering
  \scriptsize
  \setlength{\tabcolsep}{1.5pt}
  \renewcommand{\arraystretch}{0.9}
  \begin{tabular*}{\textwidth}{@{\extracolsep{\fill}}lcc|cccccc@{}}
    \toprule
    Method & ZCOCO $\uparrow$ & Avg $\uparrow$ & Ae & Hw & Pv & Sd & Th & Ve \\
    \midrule
    G-DINO~\cite{liu2024grounding}      & 47.41 & 53.15 & 45.12 & 67.54 & 58.11 & 25.84 & 70.40 & 51.87 \\
    \midrule
    ZiRa~\cite{deng2024zira}            & 44.43 & 57.63 & 39.92 & 68.00 & 64.90 & \textbf{46.26} & 77.26 & 49.47 \\
    DitHub~\cite{cappellino2025dithub} & 46.51 & \textbf{62.38} & \textbf{53.35} & 71.07 & \textbf{71.01} & 41.75 & 80.21 & \textbf{56.90} \\
    \textbf{CLA (Ours)} & \textbf{46.65}\,$\pm$\,0.19 & 61.47\,$\pm$\,0.44 & 52.64 & \textbf{71.60} & 66.20 & 41.49 & \textbf{80.46} & 56.46 \\
    \bottomrule
  
  \end{tabular*}
\end{table}

\paragraph{ODinW-O.}
Table~\ref{tab:odinwo_full} compares CLA against the baselines for which ODinW-O values are publicly reported by DitHub: G-DINO, ZiRa, and DitHub. \rev{ODinW-O is a domain-incremental regime whose tasks share classes across visually divergent domains, so one per-class module must learn a single representation of a class across these domains. CLA still attains the best ZCOCO among adapted methods and keeps Avg comparable to DitHub while trailing slightly, but here zero-shot COCO is retained rather than lifted above the unadapted Grounding DINO as on ODinW-13.} Per task, CLA remains close to DitHub on five of the six datasets, but drops markedly on Pascal VOC, by far the most data-rich of the six tasks; its heavier adaptation induces a larger drift, so the anchor trades more downstream adaptation for zero-shot retention. \rev{A per-class analysis in Appendix~\ref{sec:supp_dissociation} shows a clean dissociation on the overlapping classes: CLA preserves their COCO appearance while unanchored DitHub forgets it but fits the downstream domain better.}

\subsection{Ablation Studies}

\begin{table}[t]
  \caption{Component ablation of CLA on full-shot \textbf{ODinW-13}. $A$/$B$ denote class-specific/shared LoRA modules; aggregates are 3-seed mean$\pm$std. G-DINO is the unadapted zero-shot reference, DitHub our seed-matched reproduction.}
  \label{tab:component_ablation_odinw13_full}
  \centering
  \scriptsize
  \setlength{\tabcolsep}{1.5pt}
  \renewcommand{\arraystretch}{0.9}
  \begin{tabular*}{\textwidth}{@{\extracolsep{\fill}}lcc|ccccccccccccc@{}}
    \toprule
    Method & ZCOCO $\uparrow$ & Avg $\uparrow$ & Ae & Aq & Co & Eg & Mu & Pa & Pv & Pi & Po & Ra & Sh & Th & Ve \\
    \midrule
    G-DINO~\cite{liu2024grounding} & 47.41 & 46.80 & 19.11 & 20.82 & 64.75 & 59.98 & 25.34 & 56.27 & 54.80 & 65.94 & 22.13 & 62.02 & 32.85 & 70.38 & 57.07 \\
    DitHub~\cite{cappellino2025dithub} & $46.85{\pm}0.14$ & $61.72{\pm}0.68$ & 35.84 & 48.91 & 68.57 & 68.47 & \textbf{55.01} & 63.87 & \textbf{70.00} & 70.81 & 52.94 & 71.30 & \textbf{51.14} & 79.31 & 66.23 \\
    CLA, L2 on $A$ only & $48.10{\pm}0.06$ & $61.07{\pm}0.57$ & 34.92 & 46.83 & \textbf{72.91} & 67.98 & 52.00 & 65.23 & 64.45 & \textbf{71.09} & 52.89 & 70.79 & 50.78 & 78.55 & 65.50 \\
    CLA, L2 on $B$ only & $45.87{\pm}0.67$ & $\mathbf{62.64{\pm}0.12}$ & 37.30 & 48.49 & 72.11 & \textbf{71.52} & 52.69 & 61.99 & 69.81 & 69.41 & 54.56 & \textbf{76.64} & 50.65 & \textbf{81.55} & \textbf{67.58} \\
    CLA, L2 on $A{+}B$ & $\mathbf{48.28{\pm}0.06}$ & $62.26{\pm}0.19$ & \textbf{37.40} & \textbf{49.29} & 71.61 & 69.67 & 53.43 & \textbf{70.93} & 63.50 & 70.94 & \textbf{54.61} & 71.30 & 50.80 & 78.64 & 67.21 \\
    \bottomrule
  
  \end{tabular*}
\end{table}

\begin{table}[t]
  \caption{Component ablation of CLA on full-shot \textbf{ODinW-O}. $A$/$B$ denote class-specific/shared LoRA modules; aggregates are 3-seed mean$\pm$std. G-DINO is the unadapted zero-shot reference, DitHub our seed-matched reproduction.}
  \label{tab:component_ablation_odinwo_full}
  \centering
  \scriptsize
  \setlength{\tabcolsep}{1.5pt}
  \renewcommand{\arraystretch}{0.9}
  \begin{tabular*}{\textwidth}{@{\extracolsep{\fill}}lcc|cccccc@{}}
    \toprule
    Method & ZCOCO $\uparrow$ & Avg $\uparrow$ & Ae & Hw & Pv & Sd & Th & Ve \\
    \midrule
    G-DINO~\cite{liu2024grounding} & 47.41 & 53.15 & 45.12 & 67.54 & 58.11 & 25.84 & 70.40 & 51.87 \\
    DitHub~\cite{cappellino2025dithub} & $46.26{\pm}0.42$ & $\mathbf{62.49{\pm}0.21}$ & 52.20 & 71.11 & \textbf{70.22} & \textbf{45.06} & 80.91 & 55.42 \\
    CLA, L2 on $A$ only & $46.53{\pm}0.30$ & $61.67{\pm}0.26$ & 52.42 & \textbf{71.62} & 67.10 & 43.17 & \textbf{81.14} & 54.58 \\
    CLA, L2 on $B$ only & $46.31{\pm}0.40$ & $61.32{\pm}0.49$ & 50.26 & 70.93 & 69.66 & 42.64 & 79.80 & 54.65 \\
    CLA, L2 on $A{+}B$ & $\mathbf{46.65{\pm}0.19}$ & $61.47{\pm}0.44$ & \textbf{52.64} & 71.60 & 66.20 & 41.49 & 80.46 & \textbf{56.46} \\
    \bottomrule
  
  \end{tabular*}
\end{table}
\label{sec:ablations}

We study which LoRA components the L2 anchor should cover: the class-specific $A$ modules only, the shared $B$ only, or both. Tables~\ref{tab:component_ablation_odinw13_full} and~\ref{tab:component_ablation_odinwo_full} report these variants. The DitHub baseline in these tables is our seed-matched reproduction under the same setup, so that differences across rows isolate the effect of the anchor rather than changes in evaluation. Under this protocol CLA also exceeds DitHub on Avg, so the ZCOCO gain of Section~\ref{sec:results} does not come at the cost of adaptation.

On ODinW-13 (Table~\ref{tab:component_ablation_odinw13_full}), the two anchor targets play distinct roles. Anchoring only the class-specific $A$ modules lifts ZCOCO over DitHub at a small Avg cost, whereas anchoring only the shared $B$ keeps Avg healthy but does not preserve ZCOCO, which falls below DitHub. The zero-shot retention lever is therefore the per-class $A$: anchoring $A$ alone already recovers almost all of the ZCOCO gain. We anchor both, since the added $B$ anchor lifts Avg back above DitHub while retaining the ZCOCO gain, and we identify the class-specific $A$ modules as the primary anchoring target for retention.

On ODinW-O (Table~\ref{tab:component_ablation_odinwo_full}), where the tasks share class categories, anchoring the per-class $A$ modules alone outperforms anchoring the shared $B$ alone on both ZCOCO and Avg, while anchoring both attains the best ZCOCO at a lower Avg than the $A$ modules alone. Taken together, anchoring the per-class $A$ modules preserves zero-shot detection capability on both, whereas anchoring the shared $B$ helps downstream task adaptation in the class-incremental setting: it raises Avg on ODinW-13 but lowers it on ODinW-O, while improving ZCOCO in both. This suggests that, on the domain-incremental ODinW-O, anchoring the shared $B$ restricts the representation learning recurring classes need across divergent domains, lowering downstream adaptation while preserving zero-shot detection capability.

\subsection{Analysis}
\label{sec:analysis}

\subsubsection{A Per-Category Decomposition of the ZCOCO Gain}
\label{sec:zcoco_decomp}

To locate the ZCOCO gain, we split the $80$ COCO classes into the $21$ that have a class module in the ODinW-13 library and the $59$ that do not, which are never seen during adaptation. Table~\ref{tab:zcoco_decomp} reports AP change relative to the unadapted Grounding DINO base; the DitHub column is our seed-matched reproduction. \rev{The \textbf{CLA-selective} column is an ablation that leaves the newly introduced class modules unanchored, isolating the effect of the new-class anchor.} The per-class COCO results for all $80$ classes are reported in Appendix~\ref{sec:supp_zcoco}.

\begin{table}[!ht]
  \caption{Per-class decomposition of the ZCOCO gain on \textbf{ODinW-13} ($\Delta$AP over the unadapted Grounding DINO base; 3-seed mean$\pm$std). CLA-selective leaves the newly introduced class modules unanchored.}
  \label{tab:zcoco_decomp}
  \centering
  \scriptsize
  \setlength{\tabcolsep}{1.5pt}
  \renewcommand{\arraystretch}{0.9}
  \begin{tabular*}{\textwidth}{@{\extracolsep{\fill}}lccccc@{}}
    \toprule
    COCO classes & Number of classes & base AP & $\Delta$ DitHub & $\Delta$ CLA-selective & $\Delta$ CLA \\
    \midrule
    In ODinW-13 & $21$ & $53.41$ & $+0.59 \pm 0.10$ & $+0.58 \pm 0.32$ & $\mathbf{+1.15 \pm 0.13}$ \\
    Not in ODinW-13 & $59$ & $45.27$ & $-0.97 \pm 0.15$ & $-0.95 \pm 0.27$ & $\mathbf{+0.78 \pm 0.04}$ \\
    All COCO & $80$ & $47.41$ & $-0.56 \pm 0.14$ & $-0.55 \pm 0.28$ & $\mathbf{+0.88 \pm 0.06}$ \\
    \bottomrule
  
  \end{tabular*}
\end{table}

Both methods improve on the classes in the library, but the main difference is on the $59$ without a module: CLA preserves them while DitHub forgets them. This gap on the never-seen classes is positive and accounts for most of the ZCOCO improvement of CLA over DitHub. \rev{CLA-selective forgets these classes comparably to DitHub, so this preservation is largely lost when the new-class anchor is removed.}

\subsubsection{Per-class Library Orthogonality}
\label{sec:orthogonality}

To test whether CLA disentangles the LoRA library, we analyze the learned class-specific $A$ matrices after the full run. At each LoRA layer, we flatten and L2-normalize every $A_c$ to a unit vector $\hat{a}_c$ and form the Gram matrix $M$ of pairwise cosine similarities, $M_{cc'}=\hat{a}_c^{\top}\hat{a}_{c'}$. Table~\ref{tab:inventory} reports, averaged over LoRA layers, the \textbf{mean off-diagonal $|\cos|$} of $M$ together with the per-class backbone perturbation $\|B A_c\|$; lower values indicate stronger class separation and a smaller perturbation of the frozen detector.

\begin{table}[!ht]
  \caption{Learned class-specific $A$ library (3-seed mean$\pm$std; $|\cos|$ and $\|B A_c\|$ averaged over LoRA layers). \textbf{CLA-selective} leaves the newly introduced $A$ modules unanchored.}
  \label{tab:inventory}
  \centering
  \scriptsize
  \setlength{\tabcolsep}{1.5pt}
  \renewcommand{\arraystretch}{0.9}
  \begin{tabular*}{\textwidth}{@{\extracolsep{\fill}}lcccc|cccc@{}}
    \toprule
    & \multicolumn{4}{c|}{ODinW-13} & \multicolumn{4}{c}{ODinW-O} \\
    Method & Mean $|\cos|$ $\downarrow$ & $\|B A_c\|$ $\downarrow$ & ZCOCO $\uparrow$ & Avg $\uparrow$ & Mean $|\cos|$ $\downarrow$ & $\|B A_c\|$ $\downarrow$ & ZCOCO $\uparrow$ & Avg $\uparrow$ \\
    \midrule
    DitHub & 0.2365\,$\pm$\,0.0016 & 3.915\,$\pm$\,0.085 & 46.85\,$\pm$\,0.14 & 61.72\,$\pm$\,0.68 & 0.4368\,$\pm$\,0.0460 & 2.656\,$\pm$\,0.251 & 46.26\,$\pm$\,0.42 & \textbf{62.49}\,$\pm$\,0.21 \\
    CLA-selective & 0.2120\,$\pm$\,0.0070 & 2.583\,$\pm$\,0.199 & 46.86\,$\pm$\,0.28 & \textbf{62.81}\,$\pm$\,0.59 & 0.3397\,$\pm$\,0.0841 & 1.552\,$\pm$\,0.449 & 46.56\,$\pm$\,0.37 & 62.09\,$\pm$\,0.36 \\
    \textbf{CLA (Ours)} & \textbf{0.0501}\,$\pm$\,0.0006 & \textbf{1.235}\,$\pm$\,0.064 & \textbf{48.28}\,$\pm$\,0.06 & 62.26\,$\pm$\,0.19 & \textbf{0.1571}\,$\pm$\,0.0059 & \textbf{0.859}\,$\pm$\,0.203 & \textbf{46.65}\,$\pm$\,0.19 & 61.47\,$\pm$\,0.44 \\
    \bottomrule
  
  \end{tabular*}
\end{table}

CLA makes the learned library markedly less correlated: relative to DitHub, the mean off-diagonal cosine similarity drops on both benchmarks. This is consistent with the anchor: the class-specific $A$ modules start from a low-correlated random initialization, and CLA remains much closer to this initial low-correlation structure, whereas DitHub drifts toward a more correlated library, so the reduced correlation is a by-product of preserving the initialization rather than of any explicit decorrelation term. The same anchor yields a smaller low-rank perturbation of the pretrained detector: CLA reduces $\|B A_c\|$ by more than $3\times$ over DitHub on both benchmarks, consistent with the norm-limiting principle of ZiRa~\cite{deng2024zira}.

\paragraph{Library-average attenuation.}
\rev{This effect also has a weight-space reading. At inference, the active adapter applies the mean of the matched library factors, $B\bar A$ with $\bar A = \frac{1}{|S|}\sum_{c \in S} A_c$ over the matched class set $S$. Correlated new modules add coherently and enlarge this perturbation, whereas low-correlated ones cancel incoherently and attenuate it. Anchoring new modules near their low-correlated initialization therefore reduces interference with the pretrained detector, a weight-space explanation for the ZCOCO retention in Tables~\ref{tab:zcoco_decomp} and~\ref{tab:inventory}; we measure this attenuation over the whole library in Appendix~\ref{sec:supp_attenuation}.}

\paragraph{Isolating the role of new-class anchoring.}
\rev{The same CLA-selective variant also reshapes the learned library. As Table~\ref{tab:inventory} shows, removing the new-class anchor regresses the library toward DitHub: the mean off-diagonal $|\cos|$ rises on both benchmarks, and on ODinW-13 ZCOCO falls back to the DitHub level while Avg edges up. This regression occurs even though CLA-selective keeps $\|B A_c\|$ below DitHub, so reducing the update magnitude alone is insufficient; anchoring the new modules trades a modest amount of task plasticity for retention.}

\subsubsection{Training-Free Class Unlearning}
\label{sec:unlearning}

DitHub~\cite{cappellino2025dithub} introduces a training-free unlearning protocol: to remove class $c$, the adapted weight is modified as $W' = W - \alpha \, B A_c$ with $\alpha = 0.3$, without retraining. To examine how effectively the modules learned by CLA can be unlearned, we apply the same protocol and compare how cleanly each method separates per-class directions.

For each of the $44$ ODinW-13 classes, we apply the operation and evaluate all $13$ tasks. Table~\ref{tab:unlearn_summary} summarizes the outcome: CLA erases the target class more strongly than DitHub, while the mean absolute change on the other classes of the same task is $2.13$ for DitHub and $3.14$ for CLA. Because the perturbation CLA applies to the backbone is also far smaller, the unlearning efficiency $E$, the target AP erased per unit of perturbation, is about four times that of DitHub.

\begin{table}[!ht]
  \caption{Training-free class unlearning on \textbf{ODinW-13}, averaged over the $44$ classes.
  Target $\Delta$AP is the AP change on the task containing the removed class, so a more negative
  value is a stronger removal. Non-target $\Delta$AP is the signed mean change on the remaining
  classes against an out-of-fold reference, and non-target $|\Delta\mathrm{AP}|$ the mean absolute
  change on the other classes of the same task against the unedited library.
  $E$ is the target AP erased per unit of backbone perturbation.}
  \label{tab:unlearn_summary}
  \centering
  \scriptsize
  \setlength{\tabcolsep}{1.5pt}
  \renewcommand{\arraystretch}{0.9}
  \begin{tabular*}{\textwidth}{@{\extracolsep{\fill}}lccccc@{}}
    \toprule
    Method & Target $\Delta$AP $\downarrow$ & Non-target $\Delta$AP & Non-target $|\Delta\mathrm{AP}|$ & Perturbation $\|B A_c\|$ $\downarrow$ & Erasure per perturbation $E$ $\uparrow$ \\
    \midrule
    DitHub~\cite{cappellino2025dithub} & $-1.79$ & $\approx 0.00$ & $2.13$ & $3.915$ & $0.46$ \\
    \textbf{CLA (Ours)} & $\mathbf{-2.32}$ & $\approx 0.00$ & $3.14$ & $\mathbf{1.235}$ & $\mathbf{1.88}$ \\
    \bottomrule
  \end{tabular*}
\end{table}

\rev{Of the $44$ classes, CLA unlearns $27$ more strongly than DitHub and the rest comparably. This cleaner unlearning is consistent with the less-correlated class-specific $A$ modules learned by CLA.} Appendix~\ref{sec:supp_unlearning} reports per-class results for all $44$ classes together with qualitative examples.

\paragraph{Limitations.}
CLA targets class-incremental adaptation, where successive tasks introduce largely new classes and the goal is to retain zero-shot capability, as in ODinW-13. On the domain-incremental ODinW-O, one per-class module would need to remain aligned across the divergent domains in which a class recurs. This is possible only if a single class-wise $A$ module can represent multiple domains, but CLA does not explicitly address this domain-incremental setting. We leave this to future work.

%% file: sections/05_conclusion.tex
\section{Conclusion}

In this paper, we proposed \textbf{Class-wise LoRA Anchoring (CLA)} for incremental open-vocabulary object detection. Before each task, CLA snapshots the LoRA modules and anchors every module to that snapshot with an L2 penalty. On the class-disjoint ODinW-13, CLA outperforms both DitHub and the unadapted Grounding DINO in ZCOCO while keeping downstream accuracy comparable. Under CLA, adaptation therefore leads to a gain in zero-shot ability rather than to forgetting. This gain concentrates on COCO categories never seen during adaptation, which is consistent with CLA preserving the region-text alignment of the pretrained detector rather than overwriting it. In training-free class unlearning, the lower inter-module correlation induced by CLA also enables stronger class removal, suppressing the target class more than DitHub does.

%% file: sections/06_appendix.tex
\renewcommand{\thesection}{\Alph{section}}
\renewcommand{\thesubsection}{\Alph{section}.\arabic{subsection}}
\setcounter{table}{0}
\renewcommand{\thetable}{S\arabic{table}}
\setcounter{figure}{0}
\renewcommand{\thefigure}{S\arabic{figure}}
\setcounter{equation}{0}
\renewcommand{\theequation}{S\arabic{equation}}

\noindent
This appendix provides the per-task results deferred from the main text and extended
analyses of CLA, organized as follows.
Appendix~\ref{sec:supp_impl} details our implementation and the CLA variants.
Appendix~\ref{sec:supp_gamma} sweeps the anchor strength $\gamma_{\mathrm{a}}$ on both
ODinW-13 and ODinW-O.
Appendix~\ref{sec:app_qualitative} collects qualitative detection results and failure cases.
Appendix~\ref{sec:supp_dissociation} presents a per-class dissociation between downstream
and COCO accuracy for the shared classes on ODinW-O.
Appendix~\ref{sec:supp_zcoco} decomposes the ZCOCO gain into the COCO classes that
have a class module in the ODinW-13 library and those that do not.
Appendix~\ref{sec:supp_attenuation} reads the retained and improved ZCOCO in weight space
through library-average attenuation.
Appendix~\ref{sec:supp_unlearning} reports per-class training-free unlearning on all $44$
ODinW-13 classes, with qualitative examples and failure cases.
All notation follows the main paper; CLA denotes the L2 anchor on
both the class-specific $A$ matrices and the shared $B$ matrices, and
DitHub~\cite{cappellino2025dithub} is $\gamma_{\mathrm{a}}{=}0$. All values are 3-seed mean$\pm$std unless noted.

\section{Implementation Details}
\label{sec:supp_impl}
Our implementation is based on the public DitHub~\cite{cappellino2025dithub}
code\footnote{\url{https://github.com/aimagelab/DitHub}}: we keep the
detector, LoRA placement, rank, optimizer, learning-rate schedule, batch size, and
evaluation protocol unchanged, and CLA adds only the anchor loss
$\mathcal{L}_{\mathrm{anchor}}$ defined in the main paper. The one exception is the merge
coefficient: we use $\lambda_a{=}0.3,~\lambda_b{=}0.7$ for both ODinW-13 and ODinW-O rather
than the $\lambda_a{=}0.1$ used by DitHub for reappearing classes on ODinW-O. The anchor
strength is $\gamma_{\mathrm{a}}{=}1.0$ on ODinW-13 and $\gamma_{\mathrm{a}}{=}0.3$ on ODinW-O.
Following DitHub, the reused $A$ modules and the shared $B$ are blended with their
warm-started state at the warm-up iteration using $\lambda_a$ and $\lambda_b$; the updated
$A$ modules are then committed to the class-wise library and the shared $B$ is carried
across tasks.
CLA-selective anchors the reused class-specific $A$ modules and the shared $B$ but leaves
newly initialized $A$ modules unanchored. All reported runs use the same three random
task-order seeds for the DitHub reproduction and every CLA variant. All experiments were
conducted on a single NVIDIA RTX 6000 Ada GPU.

\section{Anchor-Strength Sweep}
\label{sec:supp_gamma}
Table~\ref{tab:supp_gamma_sweep} sweeps the anchor strength $\gamma_{\mathrm{a}}$ on
full-shot ODinW-13 and ODinW-O. On the class-disjoint ODinW-13, increasing $\gamma_{\mathrm{a}}$ consistently
improves ZCOCO and keeps Avg stable or slightly improved, with the best overall
trade-off at $\gamma_{\mathrm{a}}{=}1.0$. The class-overlapped ODinW-O instead prefers a weaker
anchor ($\gamma_{\mathrm{a}}{=}0.3$): as the ODinW-O columns
show, raising $\gamma_{\mathrm{a}}$ there trades downstream Avg for ZCOCO retention, and the
disjoint-optimal $\gamma_{\mathrm{a}}{=}1.0$ over-anchors, lowering Avg below DitHub. We
therefore adopt $\gamma_{\mathrm{a}}{=}0.3$ as the best retention--adaptation balance rather
than the ZCOCO-maximizing $\gamma_{\mathrm{a}}{=}0.7$. Thus
$\gamma_{\mathrm{a}}$ balances stability against plasticity differently on each benchmark rather
than taking a single universal value.

\begin{table}[!ht]
  \caption{Anchor-strength sweep on full-shot ODinW-13 and ODinW-O;
  $\gamma_{\mathrm{a}}{=}0$ recovers DitHub. On the class-disjoint
  ODinW-13 a larger $\gamma_{\mathrm{a}}$ improves ZCOCO while keeping Avg stable,
  whereas on the class-overlapped ODinW-O a larger $\gamma_{\mathrm{a}}$ trades Avg for ZCOCO retention; we adopt
  $\gamma_{\mathrm{a}}{=}1.0$ and $\gamma_{\mathrm{a}}{=}0.3$, respectively. The best value in
  each column is in bold.}
  \label{tab:supp_gamma_sweep}
  \centering
  \scriptsize
  \setlength{\tabcolsep}{1.5pt}
  \renewcommand{\arraystretch}{0.9}
  \begin{tabular*}{\textwidth}{@{\extracolsep{\fill}}ccc@{\hspace{14pt}}|@{\hspace{14pt}\extracolsep{0pt}}c@{\extracolsep{\fill}\hspace{2\tabcolsep}}c@{}}
    \toprule
    & \multicolumn{2}{c}{ODinW-13} & \multicolumn{2}{c}{ODinW-O} \\
    \cmidrule(l{2pt}r{30.5pt}){2-3}\cmidrule(l{2pt}r{2pt}){4-5}
    $\gamma_{\mathrm{a}}$ & Avg $\uparrow$ & ZCOCO $\uparrow$ & Avg $\uparrow$ & ZCOCO $\uparrow$ \\
    \midrule
    $0.0$ (=\,DitHub) & $61.72{\pm}0.68$ & $46.85{\pm}0.14$ & $\mathbf{62.49{\pm}0.21}$ & $46.26{\pm}0.42$ \\
    $0.3$ & $61.84{\pm}0.69$ & $48.03{\pm}0.07$ & $61.47{\pm}0.44$ & $46.65{\pm}0.19$ \\
    $0.5$ & $62.19{\pm}0.50$ & $48.13{\pm}0.13$ & $61.39{\pm}0.27$ & $46.68{\pm}0.24$ \\
    $0.7$ & $62.18{\pm}0.48$ & $48.17{\pm}0.06$ & $60.89{\pm}0.11$ & $\mathbf{46.82{\pm}0.22}$ \\
    $1.0$ & $\mathbf{62.26{\pm}0.19}$ & $\mathbf{48.28{\pm}0.06}$ & $60.94{\pm}0.46$ & $46.73{\pm}0.16$ \\
    \bottomrule
  
  \end{tabular*}
\end{table}

\section{Qualitative Detection Failure Cases}
\label{sec:app_qualitative}
\begin{figure}[htbp]
  \centering
  \includegraphics[width=\linewidth,height=0.85\textheight,keepaspectratio]{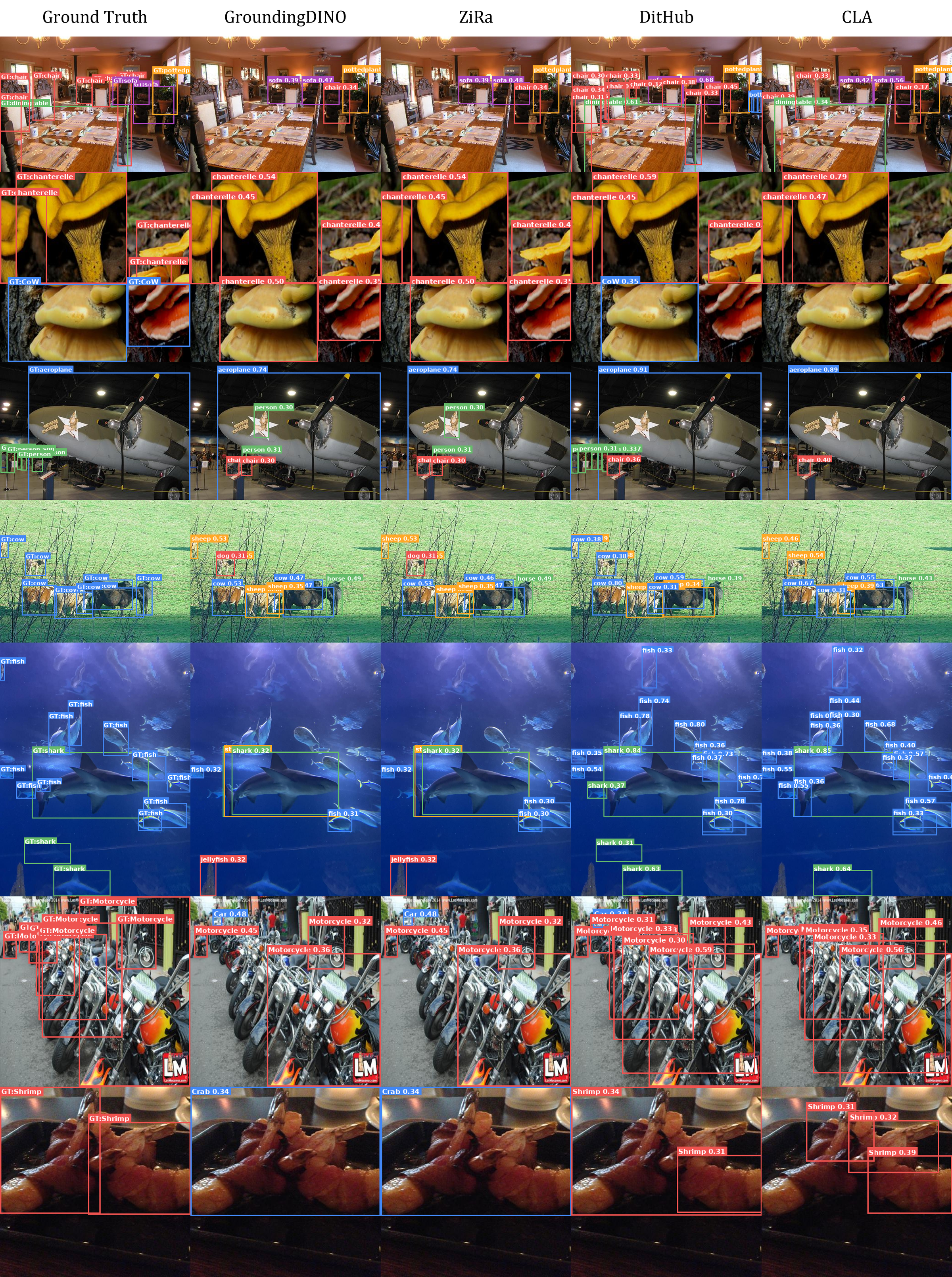}
  \caption{Qualitative failure cases on ODinW-13. Columns, left to right: ground truth,
  Grounding DINO~\cite{liu2024grounding}, ZiRa~\cite{deng2024zira},
  DitHub~\cite{cappellino2025dithub}, and CLA (Ours); predicted boxes carry confidence
  scores. Each row shows a failure case from a different task: CLA fails to detect small objects or
  confuses them with other classes, and on the North America Mushrooms task does not detect the
  Chicken-of-the-Woods (CoW) class; appearance ambiguity and clutter remain open challenges for all methods.}
  \label{fig:supp_qual_fail}
\end{figure}

Figure~\ref{fig:supp_qual_fail} shows representative ODinW-13 failure cases, using the
same column layout as Figure~\ref{fig:qualitative} in the main text.

\section{A COCO--Domain Per-Class Dissociation on ODinW-O}
\label{sec:supp_dissociation}
To analyze the Pascal VOC AP drop of CLA on ODinW-O, Table~\ref{tab:supp_dissociation}
reports, for each class that recurs in both the Pascal VOC downstream task and COCO, its AP on
the downstream task versus on COCO, for DitHub and CLA. The bold entries show a clean split:
CLA wins on COCO for every shared class, and DitHub wins on Pascal VOC for every shared class.
The anchor keeps each shared per-class module closer to its generic COCO appearance, trading a
modest amount of Pascal VOC AP for better COCO generalization. The Pascal VOC gap is
therefore better read as a domain trade-off than as CLA under-performing: DitHub specializes
each shared module to the Pascal VOC domain, while CLA holds it nearer its generic COCO
appearance. Higher Pascal VOC AP with lower COCO AP does not by itself separate specialization
from over-fitting, which would require a generalization gap measured within Pascal VOC. The
gap traces to the same anchoring mechanism that produces the ZCOCO gain of CLA.

\begin{table}[htbp]
  \caption{Per-class double dissociation on ODinW-O for the recurring shared
  classes (mean AP, DitHub vs.\ CLA); \textbf{bold} marks the higher value in each pair.
  Each column is a shared class; the two rows are the evaluation sets.}
  \label{tab:supp_dissociation}
  \centering
  \scriptsize
  \setlength{\tabcolsep}{1.5pt}
  \renewcommand{\arraystretch}{0.9}
  \begin{tabular*}{\textwidth}{@{\extracolsep{\fill}}lcccc@{}}
    \toprule
    Eval set & car & person & dog & boat \\
    \midrule
    COCO       & $48.55$ vs.\ $\mathbf{49.74}$ & $62.05$ vs.\ $\mathbf{62.60}$ & $73.85$ vs.\ $\mathbf{74.07}$ & $33.97$ vs.\ $\mathbf{35.08}$ \\
    Pascal VOC & $\mathbf{63.70}$ vs.\ $58.04$ & $\mathbf{70.68}$ vs.\ $68.35$ & $\mathbf{83.40}$ vs.\ $79.57$ & $\mathbf{63.11}$ vs.\ $58.84$ \\
    \bottomrule
  
  \end{tabular*}
\end{table}

\section{A Per-Category Decomposition of the ZCOCO Gain}
\label{sec:supp_zcoco}
Table~\ref{tab:supp_zcoco_abs} reports the \emph{absolute} post-adaptation
zero-shot COCO AP after full-shot ODinW-13, split into the $21$ COCO classes that
have a class module in the ODinW-13 library and the $59$ that do not. Head to head,
CLA improves over DitHub more on the classes outside the library than on those inside, so
the ZCOCO edge comes almost entirely from recovered performance on the classes \emph{never seen during adaptation}. CLA-selective tracks DitHub on every
group, consistent with this preservation arising from anchoring the
newly introduced class modules.

\begin{table}[htbp]
  \caption{Absolute zero-shot COCO AP after full-shot ODinW-13.
  CLA-selective leaves the newly introduced $A$ modules unanchored.
  Group means are computed at full numerical precision and can differ from the
  rounded cell values by up to $0.01$.}
  \label{tab:supp_zcoco_abs}
  \centering
  \scriptsize
  \setlength{\tabcolsep}{1.5pt}
  \renewcommand{\arraystretch}{0.9}
  \begin{tabular*}{\textwidth}{@{\extracolsep{\fill}}lccc@{}}
    \toprule
    Method & in ODinW-13 (21) & not in ODinW-13 (59) & all (80) \\
    \midrule
    G-DINO~\cite{liu2024grounding} (zero-shot) & $53.41$ & $45.27$ & $47.41$ \\
    DitHub & $54.01{\pm}0.10$ & $44.30{\pm}0.15$ & $46.85{\pm}0.14$ \\
    \textbf{CLA} & $\mathbf{54.57{\pm}0.13}$ & $\mathbf{46.05{\pm}0.04}$ & $\mathbf{48.28{\pm}0.06}$ \\
    CLA-selective & $53.99{\pm}0.32$ & $44.32{\pm}0.27$ & $46.86{\pm}0.28$ \\
    \bottomrule
  
  \end{tabular*}
\end{table}

\noindent Table~\ref{tab:supp_coco_perclass} gives the full per-class breakdown. On
the $59$ outside the library, CLA stays at or close to the base on nearly every class, whereas
DitHub drops far more, a class-by-class view consistent with
the anchor preserving the COCO knowledge DitHub forgets on these classes.

\begin{table}[htbp]
  \caption{Per-class zero-shot COCO AP after full-shot ODinW-13 adaptation
(G-DINO is the zero-shot un-adapted Grounding DINO). The $80$ COCO classes split into
the $21$ with a class module in the ODinW-13 library and the $59$ without. CLA-sel.\ is CLA-selective, which
leaves the newly introduced $A$ modules unanchored. Per class, the best value among
G-DINO, DitHub, CLA, and CLA-sel.\ is in bold, and the best among the adapted DitHub,
CLA, and CLA-sel.\ has a yellow highlight.}
  \label{tab:supp_coco_perclass}
  \centering
  \scriptsize
  \setlength{\tabcolsep}{1.5pt}
  \renewcommand{\arraystretch}{0.9}
  \begin{tabular*}{\textwidth}{@{\extracolsep{\fill}}lc|ccc@{\hskip 1em}lc|ccc@{}}
    \toprule
    Class & G-DINO & DitHub & CLA & CLA-sel. & Class & G-DINO & DitHub & CLA & CLA-sel. \\
    \midrule
\multicolumn{10}{@{}l}{\textit{In ODinW-13 (21)}}\\
\midrule
airplane & \textbf{78.4} & $76.7{\pm}0.2$ & \hi{$77.6{\pm}0.3$} & $76.7{\pm}0.3$ & bicycle & 35.7 & $37.7{\pm}0.3$ & $37.7{\pm}0.2$ & \hi{$\mathbf{38.0{\pm}1.0}$} \\
bird & 46.2 & $44.7{\pm}1.0$ & \hi{$\mathbf{47.2{\pm}0.3}$} & $44.5{\pm}0.4$ & boat & 35.6 & $34.6{\pm}0.5$ & \hi{$\mathbf{37.5{\pm}0.4}$} & $34.9{\pm}0.1$ \\
bottle & 42.7 & $46.1{\pm}0.3$ & $46.0{\pm}0.2$ & \hi{$\mathbf{46.7{\pm}0.2}$} & bus & \textbf{74.6} & $73.3{\pm}0.6$ & \hi{$74.4{\pm}0.4$} & $73.5{\pm}0.3$ \\
car & \textbf{51.3} & $48.0{\pm}0.2$ & \hi{$51.1{\pm}0.3$} & $48.6{\pm}0.6$ & cat & 80.8 & $81.6{\pm}0.2$ & \hi{$\mathbf{81.9{\pm}0.1}$} & $81.3{\pm}0.2$ \\
chair & 34.2 & \hi{$\mathbf{38.7{\pm}0.1}$} & $38.5{\pm}0.4$ & $38.5{\pm}0.2$ & couch & \textbf{53.9} & $53.2{\pm}0.4$ & \hi{$53.6{\pm}0.6$} & $52.0{\pm}0.2$ \\
cow & \textbf{66.3} & $63.5{\pm}0.4$ & \hi{$66.0{\pm}0.3$} & $63.5{\pm}0.4$ & dining table & 9.5 & \hi{$\mathbf{19.3{\pm}1.4}$} & $11.2{\pm}0.6$ & $17.3{\pm}1.6$ \\
dog & \textbf{76.2} & $74.5{\pm}0.6$ & \hi{$75.5{\pm}0.1$} & $73.9{\pm}0.1$ & horse & 66.2 & $65.4{\pm}0.8$ & \hi{$\mathbf{67.6{\pm}0.2}$} & $66.5{\pm}0.6$ \\
motorcycle & 50.7 & $52.2{\pm}0.3$ & $52.1{\pm}0.3$ & \hi{$\mathbf{52.4{\pm}0.8}$} & person & 63.3 & $61.7{\pm}0.4$ & \hi{$\mathbf{64.0{\pm}0.0}$} & $62.9{\pm}0.6$ \\
potted plant & 27.9 & \hi{$\mathbf{33.6{\pm}0.8}$} & $30.6{\pm}0.7$ & $32.3{\pm}0.6$ & sheep & 52.6 & $56.3{\pm}0.2$ & $56.2{\pm}0.8$ & \hi{$\mathbf{56.4{\pm}0.4}$} \\
train & \textbf{75.6} & $73.6{\pm}0.4$ & \hi{$74.7{\pm}0.3$} & $73.9{\pm}0.2$ & truck & 38.3 & $36.7{\pm}0.5$ & \hi{$\mathbf{39.7{\pm}0.5}$} & $37.3{\pm}1.8$ \\
tv & 61.6 & \hi{$\mathbf{62.8{\pm}0.4}$} & $62.7{\pm}0.5$ & $62.7{\pm}0.2$ &  &  &  &  &  \\
\midrule
\textbf{mean AP} & 53.41 & 54.01 & \hi{\textbf{54.57}} & 53.99 &  &  &  &  &  \\
\midrule
\multicolumn{10}{@{}l}{\textit{Not in ODinW-13 (59)}}\\
\midrule
apple & \textbf{19.1} & $18.7{\pm}0.4$ & \hi{$19.0{\pm}0.6$} & $17.9{\pm}0.4$ & backpack & 21.4 & $21.0{\pm}0.6$ & \hi{$\mathbf{21.9{\pm}0.2}$} & $21.5{\pm}0.2$ \\
banana & 26.8 & $28.0{\pm}1.1$ & \hi{$\mathbf{28.8{\pm}0.3}$} & $28.7{\pm}0.6$ & baseball bat & 43.0 & $44.2{\pm}0.8$ & \hi{$\mathbf{45.9{\pm}0.4}$} & $43.9{\pm}1.7$ \\
baseball glove & \textbf{46.6} & $44.6{\pm}0.7$ & \hi{$46.3{\pm}0.1$} & $44.5{\pm}0.3$ & bear & \textbf{78.7} & $76.8{\pm}0.3$ & \hi{$78.5{\pm}0.3$} & $76.7{\pm}0.8$ \\
bed & \textbf{49.8} & $47.8{\pm}1.9$ & \hi{$49.3{\pm}0.4$} & $46.6{\pm}0.5$ & bench & 34.3 & $33.0{\pm}0.4$ & \hi{$\mathbf{35.8{\pm}0.2}$} & $32.8{\pm}0.5$ \\
book & 12.3 & $13.4{\pm}0.8$ & \hi{$\mathbf{14.3{\pm}0.4}$} & $13.9{\pm}0.6$ & bowl & 49.9 & $49.2{\pm}0.9$ & \hi{$\mathbf{51.2{\pm}0.4}$} & $49.5{\pm}0.2$ \\
broccoli & 22.1 & $21.1{\pm}0.6$ & \hi{$\mathbf{22.5{\pm}0.1}$} & $21.5{\pm}0.9$ & cake & 43.5 & $43.2{\pm}1.0$ & \hi{$\mathbf{44.7{\pm}0.6}$} & $42.4{\pm}1.0$ \\
carrot & 19.9 & $18.8{\pm}0.3$ & \hi{$\mathbf{20.5{\pm}0.0}$} & $17.7{\pm}0.3$ & cell phone & \textbf{47.6} & $43.7{\pm}1.3$ & \hi{$\mathbf{47.6{\pm}0.3}$} & $44.5{\pm}0.8$ \\
clock & \textbf{47.3} & $43.4{\pm}0.4$ & \hi{$45.4{\pm}0.3$} & $43.0{\pm}0.4$ & cup & 51.6 & $50.4{\pm}0.4$ & \hi{$\mathbf{53.7{\pm}0.3}$} & $49.8{\pm}1.2$ \\
donut & 49.8 & $48.2{\pm}3.2$ & \hi{$\mathbf{51.5{\pm}0.5}$} & $46.9{\pm}2.8$ & elephant & 74.1 & $74.1{\pm}0.4$ & \hi{$\mathbf{74.7{\pm}0.1}$} & $74.0{\pm}0.6$ \\
fire hydrant & \textbf{76.5} & $74.7{\pm}0.5$ & \hi{$75.9{\pm}0.4$} & $73.6{\pm}0.8$ & fork & 52.9 & $50.7{\pm}0.9$ & \hi{$\mathbf{53.0{\pm}0.2}$} & $49.4{\pm}0.1$ \\
frisbee & \textbf{71.3} & $67.8{\pm}1.8$ & \hi{$70.3{\pm}0.6$} & $67.6{\pm}1.2$ & giraffe & 77.5 & $77.3{\pm}0.5$ & \hi{$\mathbf{77.7{\pm}0.3}$} & $77.2{\pm}0.3$ \\
hair drier & 10.8 & \hi{$\mathbf{13.8{\pm}1.9}$} & $13.0{\pm}0.2$ & $13.2{\pm}2.2$ & handbag & 17.8 & $17.8{\pm}0.5$ & \hi{$\mathbf{19.3{\pm}0.3}$} & $18.5{\pm}0.6$ \\
hot dog & 36.1 & $36.7{\pm}0.9$ & $36.3{\pm}0.4$ & \hi{$\mathbf{36.8{\pm}0.2}$} & keyboard & \textbf{56.5} & $54.9{\pm}0.5$ & \hi{$\mathbf{56.5{\pm}0.4}$} & $56.3{\pm}0.5$ \\
kite & \textbf{50.5} & $44.4{\pm}1.4$ & \hi{$48.9{\pm}0.1$} & $44.9{\pm}0.4$ & knife & \textbf{27.2} & $25.5{\pm}0.5$ & \hi{$27.0{\pm}0.1$} & $24.8{\pm}0.7$ \\
laptop & 75.1 & $73.1{\pm}1.3$ & \hi{$\mathbf{75.4{\pm}0.4}$} & $72.6{\pm}0.9$ & microwave & 65.4 & $64.7{\pm}0.8$ & \hi{$\mathbf{66.2{\pm}0.2}$} & $64.2{\pm}1.0$ \\
mouse & \textbf{68.3} & $63.9{\pm}0.4$ & \hi{$67.1{\pm}0.5$} & $64.1{\pm}0.9$ & orange & \textbf{35.1} & $31.4{\pm}1.9$ & \hi{$33.5{\pm}0.7$} & $28.5{\pm}1.6$ \\
oven & 19.1 & $19.1{\pm}2.2$ & \hi{$\mathbf{20.3{\pm}0.4}$} & $19.9{\pm}0.9$ & parking meter & 17.6 & $22.5{\pm}1.0$ & $22.8{\pm}1.3$ & \hi{$\mathbf{24.4{\pm}0.7}$} \\
pizza & 61.5 & $61.2{\pm}0.9$ & \hi{$\mathbf{61.7{\pm}0.3}$} & $61.5{\pm}0.6$ & refrigerator & 66.0 & $67.3{\pm}1.2$ & $67.4{\pm}0.4$ & \hi{$\mathbf{67.9{\pm}0.9}$} \\
remote & 31.1 & $31.5{\pm}2.7$ & \hi{$\mathbf{33.5{\pm}0.6}$} & $32.6{\pm}1.3$ & sandwich & 35.4 & $38.1{\pm}2.0$ & $38.0{\pm}0.1$ & \hi{$\mathbf{39.3{\pm}0.9}$} \\
scissors & 40.2 & $40.6{\pm}0.6$ & $41.4{\pm}0.8$ & \hi{$\mathbf{41.5{\pm}1.6}$} & sink & 36.3 & $36.6{\pm}2.0$ & \hi{$\mathbf{37.8{\pm}0.6}$} & $37.1{\pm}0.5$ \\
skateboard & \textbf{62.0} & $59.3{\pm}0.4$ & \hi{$61.6{\pm}0.2$} & $59.2{\pm}0.6$ & skis & \textbf{30.5} & $28.0{\pm}0.7$ & \hi{$29.5{\pm}0.2$} & $28.4{\pm}0.4$ \\
snowboard & 43.0 & $42.1{\pm}1.3$ & \hi{$\mathbf{43.3{\pm}0.9}$} & $41.6{\pm}0.5$ & spoon & \textbf{29.8} & $26.5{\pm}1.3$ & \hi{$\mathbf{29.8{\pm}0.3}$} & $27.3{\pm}0.5$ \\
sports ball & \textbf{45.4} & $38.8{\pm}1.1$ & \hi{$45.1{\pm}0.5$} & $39.8{\pm}0.4$ & stop sign & 58.0 & $56.6{\pm}2.5$ & \hi{$\mathbf{58.8{\pm}0.7}$} & $56.2{\pm}0.7$ \\
suitcase & 48.7 & $46.3{\pm}1.3$ & \hi{$\mathbf{51.3{\pm}0.4}$} & $47.2{\pm}0.9$ & surfboard & \textbf{50.0} & $47.9{\pm}0.8$ & \hi{$49.6{\pm}0.3$} & $48.9{\pm}0.7$ \\
teddy bear & 60.7 & $59.5{\pm}0.6$ & \hi{$\mathbf{61.0{\pm}0.6}$} & $59.3{\pm}0.5$ & tennis racket & \textbf{42.1} & $35.7{\pm}1.5$ & \hi{$42.0{\pm}1.7$} & $38.6{\pm}1.8$ \\
tie & \textbf{45.3} & $42.7{\pm}0.5$ & \hi{$\mathbf{45.3{\pm}0.6}$} & $43.2{\pm}1.5$ & toaster & 48.0 & \hi{$\mathbf{56.1{\pm}0.7}$} & $54.6{\pm}0.9$ & $53.0{\pm}3.1$ \\
toilet & 62.4 & $64.5{\pm}0.9$ & $64.4{\pm}0.5$ & \hi{$\mathbf{64.7{\pm}0.8}$} & toothbrush & \textbf{41.8} & $38.6{\pm}0.2$ & \hi{$41.3{\pm}1.1$} & $38.8{\pm}0.3$ \\
traffic light & 32.8 & $31.3{\pm}0.9$ & \hi{$\mathbf{33.4{\pm}0.2}$} & $30.4{\pm}1.0$ & umbrella & 43.8 & $44.8{\pm}1.1$ & \hi{$\mathbf{46.5{\pm}0.5}$} & $43.7{\pm}0.8$ \\
vase & 38.5 & $40.3{\pm}0.1$ & $40.4{\pm}0.7$ & \hi{$\mathbf{40.7{\pm}0.7}$} & wine glass & 46.3 & $46.2{\pm}0.3$ & \hi{$\mathbf{47.9{\pm}0.4}$} & $45.9{\pm}0.9$ \\
zebra & 75.6 & $75.2{\pm}0.9$ & \hi{$\mathbf{76.6{\pm}0.2}$} & $76.0{\pm}0.3$ &  &  &  &  &  \\
\midrule
\textbf{mean AP} & 45.27 & 44.30 & \hi{\textbf{46.05}} & 44.32 &  &  &  &  &  \\
    \bottomrule
  
  \end{tabular*}
\end{table}

\section{Library-Average Attenuation: A Weight-Space Reading of the Retained and Improved ZCOCO}
\label{sec:supp_attenuation}
At inference a mean of library factors $B\bar A$, with $s{=}\alpha_{\mathrm{lora}}/r{=}0.5$ the
LoRA scaling, perturbs the frozen pretrained detector and so erodes or preserves its zero-shot
detection ability. We take this mean over the whole library, $\bar A=\frac{1}{N}\sum_{c}A_c$ with
$N$ the library size, so the quantities below and Table~\ref{tab:supp_attenuation} are a
library-level diagnostic rather than a measurement of the subset matched at a particular query. This eval-time perturbation separates exactly into a per-class magnitude and an
attenuation ratio,
\begin{equation}
  \underbrace{\|sB\bar A\|}_{\text{eval-time perturbation}}
  \;=\; \underbrace{\mathrm{rms}_c\|sBA_c\|}_{\text{per-class magnitude}}
  \;\times\; \underbrace{\frac{\|sB\bar A\|}{\mathrm{rms}_c\|sBA_c\|}}_{\text{attenuation ratio }\eta}.
  \label{eq:decomp}
\end{equation}
We read the perturbation along these two axes so that a small $\|sB\bar A\|$ can be traced to
its cause: weak adaptation, that is small per-class updates, or normal-size updates that cancel
in a near-orthogonal library, and the two carry opposite implications. Anchoring lowers both factors, shrinking each per-class update and keeping the updates
near-orthogonal so that the updates cancel in the library mean, lowering the attenuation ratio $\eta$.
This second factor is fixed entirely by how the per-class updates are \emph{arranged} relative
to one another, which we make precise next. Let $G\in\mathbb{R}^{N\times N}$ be the Gram matrix of these per-class updates, with $G_{ij}=\langle sBA_i,sBA_j\rangle$,
where $N$ is the library size, the number of accumulated per-class factors
$A_c$, distinct from the number of tasks and the eval active-set size. Since
$sB\bar A=\tfrac1N\sum_c sBA_c$, both parts of $\eta$ are Gram quantities: expanding the mean into a
double sum gives $\|sB\bar A\|^2=\tfrac1{N^2}\sum_{c,c'}\langle sBA_c,sBA_{c'}\rangle=\tfrac1{N^2}\mathbf 1^\top G\mathbf 1$,
the sum of \emph{all} entries, and stacking the squared norms gives
$(\operatorname{rms}_c\|sBA_c\|)^2=\tfrac1N\sum_c\|sBA_c\|^2=\tfrac1N\operatorname{tr}(G)$, the mean diagonal, so
\begin{equation}
\eta^2
  =\frac{\|sB\bar A\|^2}{(\operatorname{rms}_c\|sBA_c\|)^2}
  =\frac{\mathbf 1^\top G\mathbf 1}{N\operatorname{tr}(G)}
  =\frac1N\Big(1+\frac{\sum_{i\ne j}G_{ij}}{\operatorname{tr}(G)}\Big).
  \label{eq:eta-gram}
\end{equation}
Equivalently, \eqref{eq:decomp} refines into a three-factor form
\begin{equation}
\|sB\bar A\|
  =\operatorname{rms}_c\|sBA_c\|
   \times\underbrace{\tfrac1{\sqrt N}}_{\text{averaging gain}}
   \times\underbrace{\sqrt{1+\tfrac{\sum_{i\ne j}G_{ij}}{\operatorname{tr}(G)}}}_{\text{off-diagonal penalty}},
  \label{eq:three}
\end{equation}
whose last factor equals $1$ iff the off-diagonal entries of $G$ sum to zero, as they do in
particular for an orthogonal library, which gives $\eta=1/\sqrt N$. It rises to $\sqrt N$
(no cancellation, $\eta=1$) as the off-diagonal mass reaches $(N{-}1)\operatorname{tr}(G)$, and
drops below $1$ when that mass is negative, so $1/\sqrt N$ bounds $\eta$ from below only once
the off-diagonal mass is non-negative. Writing the inter-class correlation
as the mean off-diagonal over the mean diagonal of $G$,
\begin{equation}
\bar\rho:=\frac{\sum_{i\ne j}G_{ij}}{(N-1)\operatorname{tr}(G)}
        =\frac{\text{mean off-diagonal of }G}{\text{mean diagonal of }G},
  \label{eq:rho}
\end{equation}
the attenuation is a strictly increasing function of the library orthogonality alone,
\begin{equation}
\eta=\Phi(\bar\rho):=\sqrt{\frac{1+(N-1)\bar\rho}{N}},
  \qquad \Phi'(\bar\rho)=\frac{N-1}{2N\,\eta}>0,
  \label{eq:phi}
\end{equation}
ranging over the admissible $\bar\rho\in[-\tfrac{1}{N-1},1]$ from $\eta{=}0$ (updates cancelling
exactly) through $1/\sqrt N$ (orthogonal, $\bar\rho{=}0$) to $1$ (aligned, $\bar\rho{=}1$). Every
library we measure has $\bar\rho>0$, so $1/\sqrt N$ is the effective floor in this regime.
Note that $\bar\rho$ is a signed quantity on the effective updates $sBA_c$, whereas the mean
$|\cos|$ of Table~\ref{tab:inventory} is unsigned and computed on the factors $A_c$ alone:
only a signed quantity can register the cancellation that $\eta$ measures, so the two describe
related but different geometries. The
eval-time perturbation drop is therefore \emph{not} an independent lever but a direct
consequence of the near-orthogonal geometry anchoring maintains; we report $\eta$ and the
resulting $\|sB\bar A\|$ in Table~\ref{tab:supp_attenuation}.

Both benchmarks show the same ordering: the per-class updates of CLA cancel more at inference,
giving a smaller $\eta$ and a smaller eval-time perturbation, consistent with the higher retained
ZCOCO. On ODinW-13 the eval gap of about $6.5\times$ factors into a $3.33\times$ per-class-magnitude
term and a $1.98\times$ cancellation term $\eta$, so anchoring shrinks both, but only $\eta$ is the
library-cancellation effect. The small ODinW-O library, with $N{=}5$, raises the orthogonal value
$1/\sqrt N$ and caps the attainable cancellation.

\begin{table}[htbp]
  \caption{Library-average attenuation on the adapted linear layers, weight-only,
  no inference. $\eta$~\eqref{eq:phi} is the attenuation ratio, set by the inter-class
  correlation $\bar\rho$~\eqref{eq:rho} (the Gram off-diagonal over diagonal). \emph{init} is the random-init weights.}
  \label{tab:supp_attenuation}
  \centering
  \scriptsize
  \setlength{\tabcolsep}{1.5pt}
  \renewcommand{\arraystretch}{0.9}
  \begin{tabular*}{\textwidth}{@{\extracolsep{\fill}}lcccc@{}}
    \toprule
    & \multicolumn{1}{c}{per-class magnitude} & \multicolumn{1}{c}{attenuation ratio} & \multicolumn{1}{c}{eval-time perturbation} & \\
    \cmidrule(lr){2-2}\cmidrule(lr){3-3}\cmidrule(lr){4-4}
    Method & $\|sBA_c\|$ & $\eta$ & $\|sB\bar A\|$ & ZCOCO \\
    \midrule
    \multicolumn{5}{@{}l}{\textit{ODinW-13 (class-incremental, $N{=}44$)}} \\
    init (random) & -- & -- & -- & $47.41$ \\
    DitHub & $4.23{\pm}0.11$ & $0.588{\pm}0.002$ & $2.49{\pm}0.07$ & $46.85{\pm}0.14$ \\
    \textbf{CLA} & $\mathbf{1.27{\pm}0.08}$ & $\mathbf{0.297{\pm}0.004}$ & $\mathbf{0.38{\pm}0.02}$ & $\mathbf{48.28{\pm}0.06}$ \\
    \midrule
    \multicolumn{5}{@{}l}{\textit{ODinW-O (domain-incremental, small library $N{=}5$)}} \\
    init (random) & -- & -- & -- & $47.41$ \\
    DitHub & $2.72{\pm}0.29$ & $0.753{\pm}0.036$ & $2.05{\pm}0.31$ & $46.26{\pm}0.42$ \\
    \textbf{CLA} & $\mathbf{0.94{\pm}0.30}$ & $\mathbf{0.615{\pm}0.003}$ & $\mathbf{0.58{\pm}0.18}$ & $\mathbf{46.65{\pm}0.19}$ \\
    \bottomrule
  
  \end{tabular*}
\end{table}

\section{Per-Class and Qualitative Unlearning Results}
\label{sec:supp_unlearning}

\begin{figure}[htbp]
  \centering
  \includegraphics[width=\linewidth]{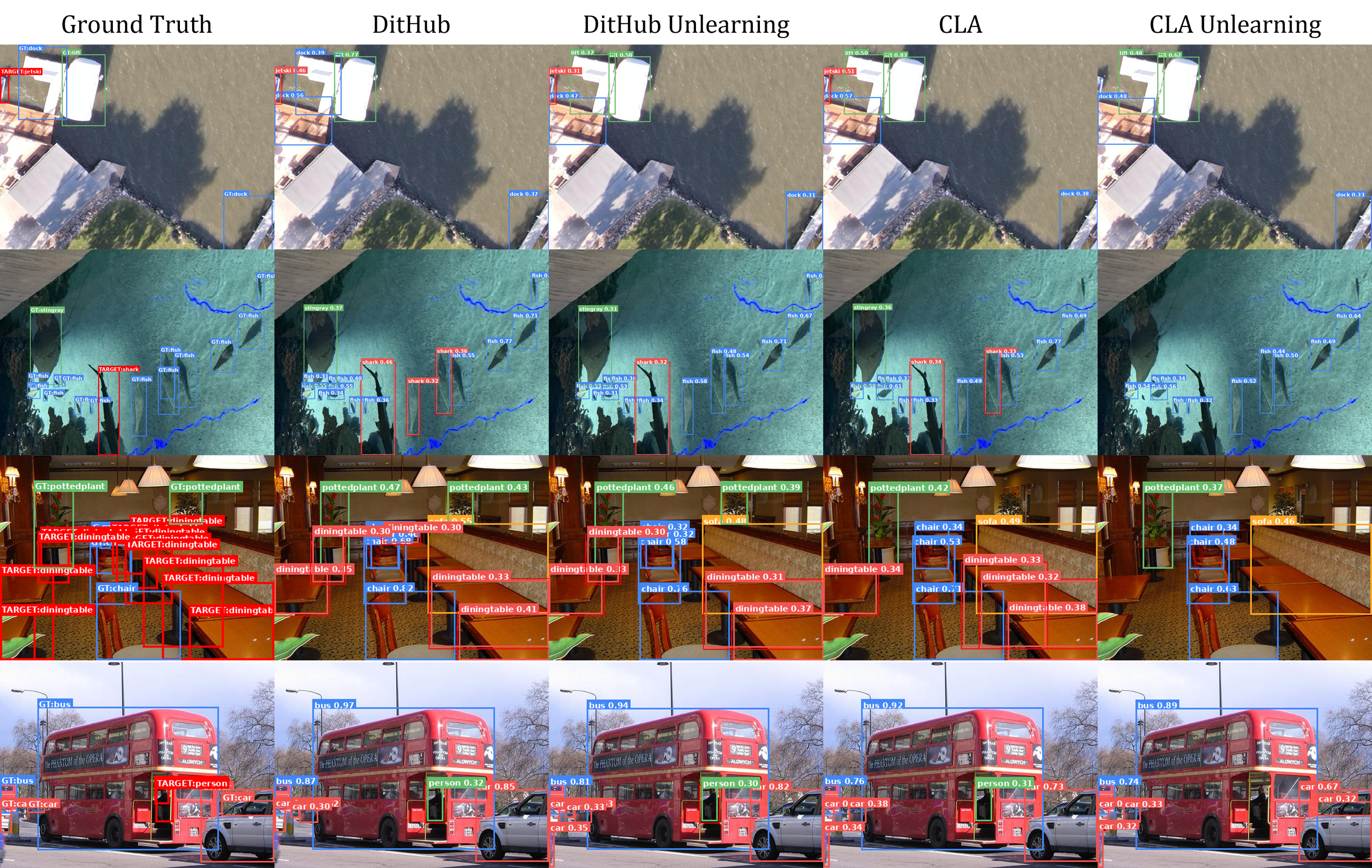}
  \caption{Qualitative class unlearning results on ODinW-13. Columns: ground truth (target class in red), DitHub, DitHub after unlearning, CLA, and CLA after unlearning ($W'{=}W-\alpha B A_c$). CLA erases the target more thoroughly while sparing other detections.}
  \label{fig:unlearning}
\end{figure}

\begin{figure}[htbp]
  \centering
  \includegraphics[width=\linewidth,height=0.85\textheight,keepaspectratio]{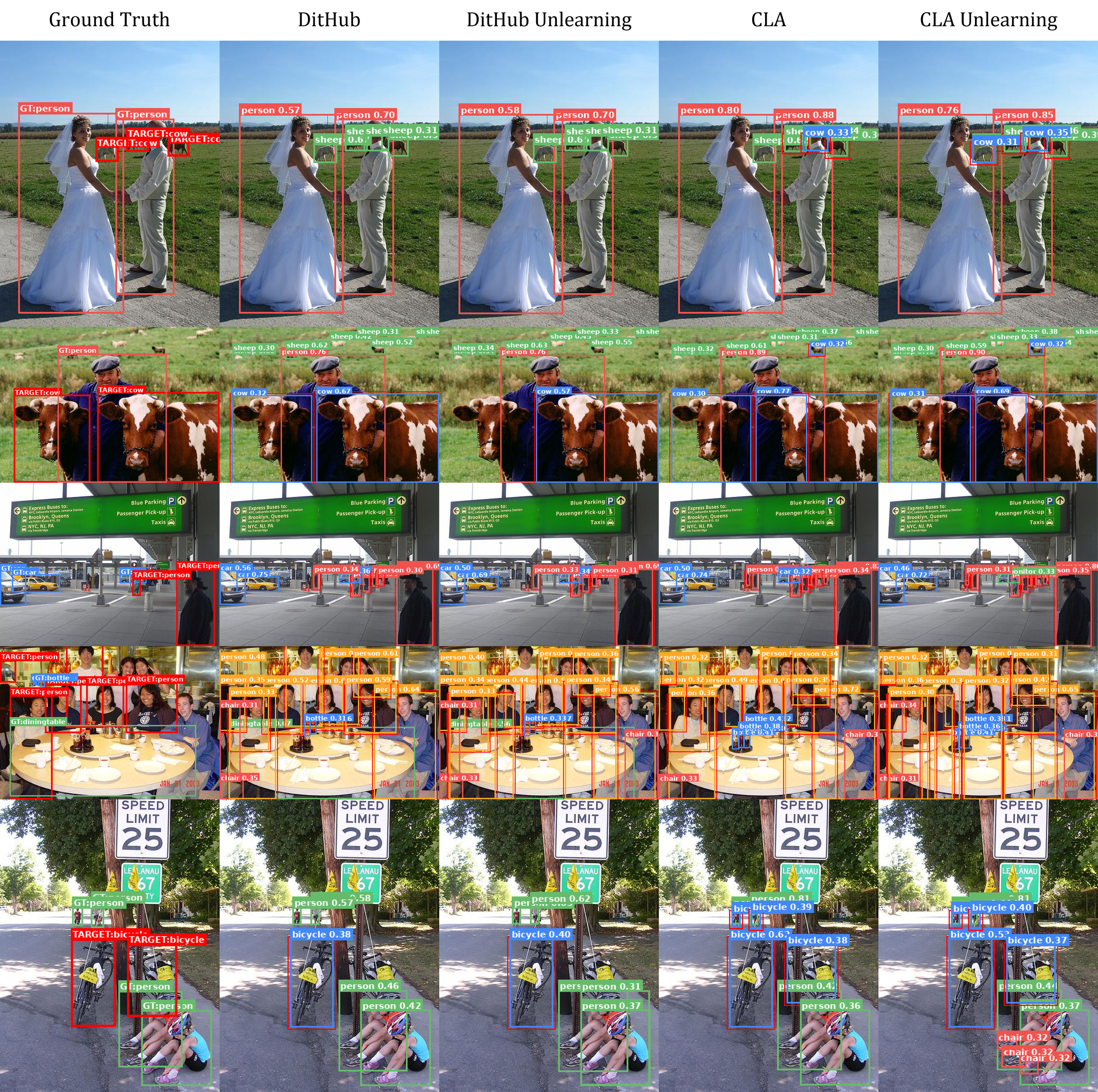}
  \caption{Qualitative class-unlearning failure cases on ODinW-13. Columns: ground truth
  (target class in red), DitHub, DitHub after unlearning, CLA, and CLA after unlearning
  ($W'{=}W-\alpha B A_c$). In some cases the unlearned target class is still detected after the
  edit, or a different class becomes newly detected. These unlearning failure cases, seen for both
  DitHub and CLA, show that a subtractive class module edit does not completely eliminate
  the detection ability for the target class.}
  \label{fig:supp_unlearn_fail}
\end{figure}

\noindent Figure~\ref{fig:unlearning} shows qualitative unlearning examples on ODinW-13, and
Figure~\ref{fig:supp_unlearn_fail} representative failures, where the target class survives the
edit or a new class becomes detected.

We apply the training-free unlearning operation $W'{=}W-\alpha B A_c$ with
$\alpha{=}0.3$ to each of the $44$ ODinW-13 classes and measure the resulting change in
the AP of the task that contains $c$,
\begin{equation}
  \Delta_{\mathrm{tgt}}(c) = \text{target\_AP}(c) - \text{baseline\_AP}(c),
\end{equation}
where $\text{target\_AP}(c)$ removes the module $A_c$ and $\text{baseline\_AP}(c)$ is an
out-of-fold estimate with $A_c$ intact (the mean over same-method runs that remove a different
class); a more negative $\Delta_{\mathrm{tgt}}$ is a stronger removal. Table~\ref{tab:supp_unlearn_full}
reports $\Delta_{\mathrm{tgt}}$ for all $44$ classes and the 44-class mean: CLA removes the target
more strongly than DitHub on average and on most classes. The collateral $\Delta_{\mathrm{non}}$,
the mean AP change on the non-target classes when $A_c$ is removed, stays near zero for both
methods. Measured against the unedited library, the other classes of the same task change by
$2.13$ (DitHub) and $3.14$ (CLA) on average. The advantage is largest on the seven Aquarium marine species and the
package class, where subtracting $B A_c$ suppresses the target more strongly, whereas on the
remaining classes CLA is marginally weaker.

\begin{table}[htbp]
  \caption{Per-class training-free unlearning on all $44$ ODinW-13 classes:
  target-task AP drop $\Delta_{\mathrm{tgt}}$ (more negative is
  stronger removal); the bottom row gives the 44-class mean and the collateral
  $\Delta_{\mathrm{non}}$ on non-target classes.
  CoW is the Chicken-of-the-Woods mushroom (North America Mushrooms), distinct from
  the animal cow. A yellow highlight marks CLA cells that unlearn the target more strongly than
  DitHub; a blue highlight marks crab and lobster, whose positive $\Delta_{\mathrm{tgt}}$ is analyzed below.
  Bold marks the seven Aquarium marine species, the classes CLA unlearns most strongly.}
  \label{tab:supp_unlearn_full}
  \centering
  \scriptsize
  \setlength{\tabcolsep}{1.5pt}
  \renewcommand{\arraystretch}{0.9}
  \begin{tabular*}{\textwidth}{@{\extracolsep{\fill}}lcc@{\hskip 1.2em}lcc@{}}
    \toprule
    Class & DitHub & CLA & Class & DitHub & CLA \\
    \midrule
    aeroplane & $-2.15{\pm}0.10$ & $-1.98{\pm}0.34$ & jetski & $-3.05{\pm}1.02$ & \hi{$-3.08{\pm}0.66$} \\
    ambulance & $-1.50{\pm}0.29$ & $-1.32{\pm}0.00$ & lift & $-3.18{\pm}0.90$ & $-3.17{\pm}0.59$ \\
    bicycle & $-1.51{\pm}0.17$ & $-1.48{\pm}0.24$ & lobster & $-0.25{\pm}0.53$ & \hipos{$+0.73{\pm}0.75$} \\
    bird & $-1.60{\pm}0.15$ & $-1.53{\pm}0.26$ & motorbike & $-1.50{\pm}0.14$ & $-1.16{\pm}0.23$ \\
    boat & $-1.27{\pm}0.14$ & \hi{$-2.19{\pm}0.78$} & motorcycle & $-1.11{\pm}0.38$ & $-0.71{\pm}0.39$ \\
    bottle & $-1.48{\pm}0.14$ & \hi{$-1.59{\pm}0.11$} & package & $+0.33{\pm}1.01$ & \hi{$-3.34{\pm}1.26$} \\
    bus & $-0.63{\pm}0.23$ & \hi{$-1.33{\pm}0.35$} & penguin & $-3.70{\pm}0.79$ & \hi{$\mathbf{-5.98{\pm}0.36}$} \\
    car & $-0.54{\pm}0.16$ & \hi{$-0.86{\pm}0.13$} & person & $-0.87{\pm}0.45$ & \hi{$-1.54{\pm}0.31$} \\
    cat & $-1.71{\pm}0.19$ & \hi{$-2.28{\pm}0.13$} & pistol & $-0.12{\pm}0.26$ & \hi{$-0.24{\pm}0.49$} \\
    chair & $-1.39{\pm}0.13$ & \hi{$-2.33{\pm}0.16$} & pothole & $-2.64{\pm}1.28$ & $-2.31{\pm}1.45$ \\
    chanterelle & $-2.35{\pm}0.95$ & \hi{$-2.89{\pm}0.36$} & pottedplant & $-1.52{\pm}0.15$ & \hi{$-1.55{\pm}0.08$} \\
    CoW & $-1.39{\pm}0.55$ & \hi{$-1.69{\pm}0.40$} & puffin & $-3.70{\pm}0.55$ & \hi{$\mathbf{-5.27{\pm}1.17}$} \\
    cottontail-Rabbit & $-0.26{\pm}1.83$ & \hi{$-1.22{\pm}2.25$} & raccoon & $-1.37{\pm}0.32$ & \hi{$-1.62{\pm}0.97$} \\
    cow & $-0.81{\pm}0.08$ & $-0.77{\pm}0.16$ & shark & $-3.90{\pm}0.79$ & \hi{$\mathbf{-6.17{\pm}0.69}$} \\
    crab & $-0.47{\pm}0.98$ & \hipos{$+0.28{\pm}1.49$} & sheep & $-1.65{\pm}0.12$ & $-1.22{\pm}0.36$ \\
    diningtable & $-1.56{\pm}0.14$ & $-1.34{\pm}0.13$ & shrimp & $-1.14{\pm}1.07$ & \hi{$-1.65{\pm}2.15$} \\
    dock & $-2.97{\pm}1.11$ & \hi{$-3.13{\pm}0.82$} & sofa & $-1.61{\pm}0.13$ & \hi{$-1.71{\pm}0.18$} \\
    dog & $-0.77{\pm}0.10$ & \hi{$-1.16{\pm}0.27$} & starfish & $-3.66{\pm}0.54$ & \hi{$\mathbf{-4.93{\pm}0.93}$} \\
    fish & $-3.68{\pm}0.76$ & \hi{$\mathbf{-7.86{\pm}0.26}$} & stingray & $-3.81{\pm}0.81$ & \hi{$\mathbf{-6.76{\pm}0.54}$} \\
    hand & $-2.57{\pm}0.40$ & $-2.37{\pm}0.33$ & train & $-1.72{\pm}0.16$ & $-1.54{\pm}0.32$ \\
    horse & $-1.53{\pm}0.14$ & $-1.32{\pm}0.02$ & truck & $-1.40{\pm}0.41$ & $-1.25{\pm}0.29$ \\
    jellyfish & $-3.73{\pm}0.84$ & \hi{$\mathbf{-5.43{\pm}1.08}$} & tvmonitor & $-1.52{\pm}0.15$ & \hi{$-1.62{\pm}0.08$} \\
    \midrule
    mean $\Delta_{\mathrm{tgt}}$ (44) & $-1.79$ & \hi{$-2.32$} & \multicolumn{3}{r@{}}{CLA stronger on $27/44$ classes;\quad $\Delta_{\mathrm{non}}\approx 0.00$ (both)} \\
    \bottomrule
  
  \end{tabular*}
\end{table}

\begin{table}[htbp]
  \caption{Intra-task confusion on the ShellfishOpenImages task (mean AP).
  Each row removes one class module (training-free) and reports the per-class AP, the
  task mean AP, and the change $\Delta_{\mathrm{tgt}}$ from the intact first row. This
  within-task $\Delta_{\mathrm{tgt}}$ corresponds to the out-of-fold $\Delta_{\mathrm{tgt}}$
  in Table~\ref{tab:supp_unlearn_full} up to the baseline convention. In the removal rows,
  \textcolor{green!55!black}{$\blacktriangle$}/\textcolor{red!75!black}{$\blacktriangledown$}
  mark each per-class AP rising/falling versus the intact row.}
  \label{tab:supp_shellfish}
  \centering
  \scriptsize
  \setlength{\tabcolsep}{1.5pt}
  \renewcommand{\arraystretch}{0.9}
  \begin{tabular*}{\textwidth}{@{\extracolsep{\fill}}l|ccccc@{}}
    \toprule
    Unlearned & crab & lobster & shrimp & task mean AP & $\Delta_{\mathrm{tgt}}$ \\
    \midrule
    None (intact) & $53.02$ & $52.54$ & $46.83$ & $50.80$ & $0.00$ \\
    crab    & \tdn{$50.74$} & \tdn{$51.18$} & \tup{$49.05$} & $50.32$ & $-0.48$ \\
    lobster & \tdn{$49.59$} & \tup{$54.18$} & \tup{$48.54$} & $50.77$ & $-0.03$ \\
    shrimp  & \tdn{$47.65$} & \tdn{$49.87$} & \tup{$47.65$} & $48.39$ & $-2.41$ \\
    \bottomrule
  
  \end{tabular*}
\end{table}

\paragraph{A limitation: positive $\Delta_{\mathrm{tgt}}$ on the small ShellfishOpenImages task.}
\rev{Removing a class module should lower the AP of that class. On ShellfishOpenImages, a small
task with three visually similar, co-occurring classes where the crab, lobster, and shrimp
categories have only $19$, $31$, and $66$ ground-truth boxes, unlearning fails on these classes
(Table~\ref{tab:supp_shellfish}):
erasing the lobster or shrimp module \emph{raises} the AP of that same class, while crab, the
minority class, falls whenever any module is removed. A module that can be deleted without
lowering, or while even raising, the AP of the class the module represents was not contributing a
useful class-specific representation. When the classes are few and share features, the anchor
holds each class module close to initialization and can keep the module from learning a
sufficiently distinct representation, so removing the module does not erase the class. Unlearning
here has a side-effect: removing a representation that mostly confuses a class with the
co-occurring siblings can even raise the detection performance of that class rather than erase the
class.}

%% file: checklist.tex
\section*{NeurIPS Paper Checklist}

\begin{enumerate}

\item {\bf Claims}
    \item[] Question: Do the main claims made in the abstract and introduction accurately reflect the paper's contributions and scope?
    \item[] Answer: \answerYes{}.
    \item[] Justification: The abstract and Section~\ref{sec:intro} state the three contributions: the class-wise anchor, the zero-shot retention it produces, and the library analysis that accounts for it. The claims are scoped to the class-incremental setting, and the weaker domain-incremental behaviour is stated rather than omitted.

\item {\bf Limitations}
    \item[] Question: Does the paper discuss the limitations of the work performed by the authors?
    \item[] Answer: \answerYes{}.
    \item[] Justification: The Limitations paragraph at the end of Section~\ref{sec:analysis} states that CLA targets class-incremental adaptation and does not address the domain-incremental setting, in which one per-class module must stay aligned across divergent domains. Section~\ref{sec:ablations} and Appendix~\ref{sec:supp_gamma} report the settings in which the anchor costs downstream task accuracy, and Appendix~\ref{sec:supp_unlearning} reports classes for which the unlearning edit fails.

\item {\bf Theory assumptions and proofs}
    \item[] Question: For each theoretical result, does the paper provide the full set of assumptions and a complete (and correct) proof?
    \item[] Answer: \answerYes{}.
    \item[] Justification: The paper proves no theorems. Its only analytical result, the decomposition of the eval-time perturbation into a per-class magnitude and an attenuation ratio, is derived in full from the stated definitions in Appendix~\ref{sec:supp_attenuation}.

    \item {\bf Experimental result reproducibility}
    \item[] Question: Does the paper fully disclose all the information needed to reproduce the main experimental results of the paper to the extent that it affects the main claims and/or conclusions of the paper (regardless of whether the code and data are provided or not)?
    \item[] Answer: \answerYes{}.
    \item[] Justification: Section~\ref{sec:cla} and Algorithm~\ref{alg:cla} define the anchor and the per-task training loop. Section~\ref{sec:evaluation} and Appendix~\ref{sec:supp_impl} give the base detector, benchmarks, metrics, merge coefficients, anchor strength, and iterations per seed. Both benchmarks derive from the public ODinW suite and retention is measured on COCO val2017.

\item {\bf Open access to data and code}
    \item[] Question: Does the paper provide open access to the data and code, with sufficient instructions to faithfully reproduce the main experimental results, as described in supplemental material?
    \item[] Answer: \codeanswer{}.
    \item[] Justification: The benchmarks are public (ODinW, COCO) and our implementation modifies the public DitHub code base, credited with its repository in Appendix~\ref{sec:supp_impl}; CLA adds only the anchor loss, which Section~\ref{sec:cla} and Algorithm~\ref{alg:cla} specify in full, together with every hyperparameter we changed. \codejustification

\item {\bf Experimental setting/details}
    \item[] Question: Does the paper specify all the training and test details (e.g., data splits, hyperparameters, how they were chosen, type of optimizer) necessary to understand the results?
    \item[] Answer: \answerYes{}.
    \item[] Justification: Section~\ref{sec:evaluation} reports the base detector, the two benchmarks, the metrics, the merge coefficients $\lambda_a$ and $\lambda_b$, the anchor strength $\gamma_{\mathrm{a}}$, and the number of iterations per seed. Appendix~\ref{sec:supp_impl} states which settings are inherited unchanged from DitHub and which we altered, and Appendix~\ref{sec:supp_gamma} reports the sweep from which $\gamma_{\mathrm{a}}$ was selected.

\item {\bf Experiment statistical significance}
    \item[] Question: Does the paper report error bars suitably and correctly defined or other appropriate information about the statistical significance of the experiments?
    \item[] Answer: \answerYes{}.
    \item[] Justification: Every number from our own runs is a mean $\pm$ sample standard deviation over three random task-order seeds, as stated in Section~\ref{sec:evaluation}. Baseline rows quoted from the DitHub paper carry no error bars; the table captions mark which rows are quoted and which are our seed-matched reproductions, so the two are never compared as if equally resolved.

\item {\bf Experiments compute resources}
    \item[] Question: For each experiment, does the paper provide sufficient information on the computer resources (type of compute workers, memory, time of execution) needed to reproduce the experiments?
    \item[] Answer: \answerYes{}.
    \item[] Justification: Appendix~\ref{sec:supp_impl} states that all experiments ran on a single NVIDIA RTX 6000 Ada GPU, and Section~\ref{sec:evaluation} gives the number of training iterations per seed for each benchmark. Total wall-clock time is not reported.
    
\item {\bf Code of ethics}
    \item[] Question: Does the research conducted in the paper conform, in every respect, with the NeurIPS Code of Ethics \url{https://neurips.cc/public/EthicsGuidelines}?
    \item[] Answer: \answerYes{}.
    \item[] Justification: The research conforms to the NeurIPS Code of Ethics. It uses only public benchmarks and involves no human subjects and no personally identifying data.

\item {\bf Broader impacts}
    \item[] Question: Does the paper discuss both potential positive societal impacts and negative societal impacts of the work performed?
    \item[] Answer: \answerNA{}.
    \item[] Justification: The paper studies a training-time regularizer for continual adaptation and evaluates it on public detection benchmarks. It introduces no dataset, deployed system, or application, so it carries no societal impact beyond that of open-vocabulary detection in general.
    
\item {\bf Safeguards}
    \item[] Question: Does the paper describe safeguards that have been put in place for responsible release of data or models that have a high risk for misuse (e.g., pre-trained language models, image generators, or scraped datasets)?
    \item[] Answer: \answerNA{}.
    \item[] Justification: The paper releases no dataset and no pretrained model. It uses publicly released detectors and benchmarks only.

\item {\bf Licenses for existing assets}
    \item[] Question: Are the creators or original owners of assets (e.g., code, data, models), used in the paper, properly credited and are the license and terms of use explicitly mentioned and properly respected?
    \item[] Answer: \answerYes{}.
    \item[] Justification: Grounding DINO, DitHub, ZiRa, ODinW, and COCO are cited where they are used, and Appendix~\ref{sec:supp_impl} links the public DitHub repository on which our implementation is based.

\item {\bf New assets}
    \item[] Question: Are new assets introduced in the paper well documented and is the documentation provided alongside the assets?
    \item[] Answer: \answerNA{}.
    \item[] Justification: The paper introduces no new assets.

\item {\bf Crowdsourcing and research with human subjects}
    \item[] Question: For crowdsourcing experiments and research with human subjects, does the paper include the full text of instructions given to participants and screenshots, if applicable, as well as details about compensation (if any)? 
    \item[] Answer: \answerNA{}.
    \item[] Justification: The paper involves no crowdsourcing and no research with human subjects.

\item {\bf Institutional review board (IRB) approvals or equivalent for research with human subjects}
    \item[] Question: Does the paper describe potential risks incurred by study participants, whether such risks were disclosed to the subjects, and whether Institutional Review Board (IRB) approvals (or an equivalent approval/review based on the requirements of your country or institution) were obtained?
    \item[] Answer: \answerNA{}.
    \item[] Justification: The paper involves no research with human subjects.

\item {\bf Declaration of LLM usage}
    \item[] Question: Does the paper describe the usage of LLMs if it is an important, original, or non-standard component of the core methods in this research? Note that if the LLM is used only for writing, editing, or formatting purposes and does \emph{not} impact the core methodology, scientific rigor, or originality of the research, declaration is not required.
    \item[] Answer: \answerNA{}.
    \item[] Justification: The base detector contains a text encoder, but no large language model is an important, original, or non-standard component of the proposed method.

\end{enumerate}